\documentclass[11pt]{article}

\usepackage[final]{acl}
\usepackage{booktabs}
\usepackage[table]{xcolor}
\usepackage{amsmath}
\usepackage{multirow}
\usepackage{listings}

\usepackage{graphicx}
\usepackage{subcaption}

\DeclareCaptionStyle{ruled}{labelfont=normalfont,labelsep=colon,strut=off} 
\usepackage{times}
\usepackage{latexsym}

\usepackage[T1]{fontenc}

\usepackage[utf8]{inputenc}

\usepackage{microtype}

\usepackage{inconsolata}

\usepackage{graphicx}

\title{FAQ: Mitigating Quantization Error via Regenerating Calibration Data with Family-Aware Quantization}

\author{
 \textbf{Haiyang Xiao\textsuperscript{1,*}},
 \textbf{Weiqing Li\textsuperscript{1,*}},
 \textbf{Jinyue Guo\textsuperscript{1,2}},
 \textbf{Guochao Jiang\textsuperscript{1}},
 \textbf{Guohua Liu\textsuperscript{1}},
 \textbf{Yuewei Zhang\textsuperscript{1,$\dag$}}
\\
\\
 \textsuperscript{1}Alibaba Cloud Computing,
 \textsuperscript{2}Chinese Academy of Sciences
\\
   \texttt{\{xiaohaiyang.xhy, liweiqing.lwq\}@alibaba-inc.com} \\
   \texttt{liyou.zyw@alibaba-inc.com}
}

\begin{document}
\maketitle
\def\thefootnote{*}\footnotetext{Equal contributions.}
\def\thefootnote{$\dag$}\footnotetext{Corresponding author.}
\def\thefootnote{\arabic{footnote}}
\begin{abstract}

Although post-training quantization (PTQ) provides an efficient numerical compression scheme for deploying large language models (LLMs) on resource-constrained devices, the representativeness and universality of calibration data remain a core bottleneck in determining the accuracy of quantization parameters. Traditional PTQ methods typically rely on limited samples, making it difficult to capture the activation distribution during the inference phase, leading to biases in quantization parameters. To address this, we propose \textbf{FAQ} (Family-Aware Quantization), a calibration data regeneration framework that leverages prior knowledge from LLMs of the same family to generate high-fidelity calibration samples. Specifically, FAQ first inputs the original calibration samples into a larger LLM from the same family as the target model, regenerating a series of high-fidelity calibration data using a highly consistent knowledge system. Subsequently, this data, carrying Chain-of-Thought reasoning and conforming to the expected activation distribution, undergoes group competition under expert guidance to select the best samples, which are then re-normalized to enhance the effectiveness of standard PTQ. Experiments on multiple model series, including Qwen3-8B, show that FAQ reduces accuracy loss by up to 28.5\% compared to the baseline with original calibration data, demonstrating its powerful potential and contribution.

\end{abstract}

\section{Introduction}




The substantial computational and memory demands of large language models (LLMs) continue to drive inference optimization across algorithmic approaches and model-level techniques~\cite{jiang2025flashthinkearlyexitmethod, quan-etal-2025-rasd}, with weight quantization playing a central role. Post-training quantization (PTQ) converts pre-trained weights and activations from high-precision formats (e.g., FP16) to low-bit integers (e.g., INT8/INT4) without retraining, substantially reducing memory footprints and accelerating inference on resource-constrained hardware.

\begin{figure}[t]
\centering
\includegraphics[width=0.9\columnwidth]{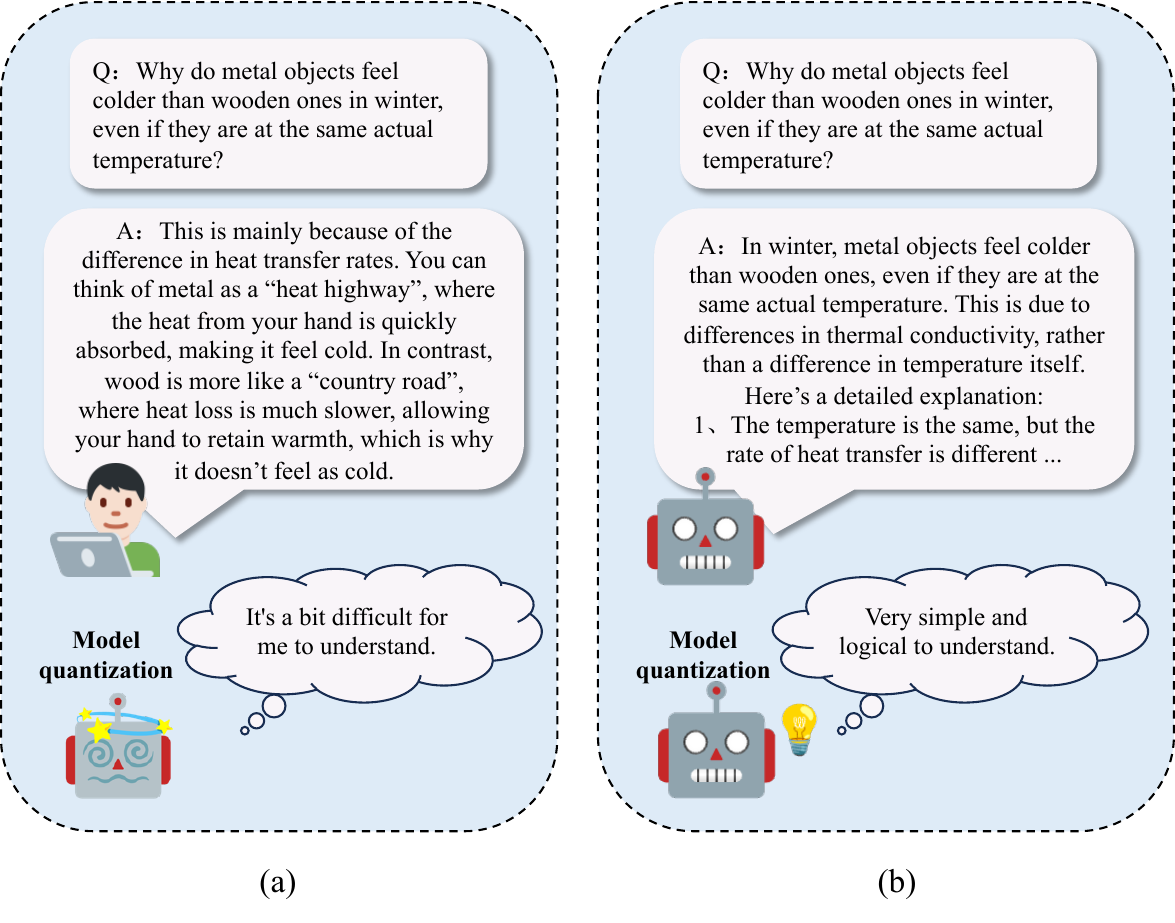} 
\caption{The impact of calibration data on quantization: (a) Traditional PTQ relies on human-provided calibration data, which may not align well with the model's internal activation patterns, leading to suboptimal quantization. (b) FAQ leverages a larger in-family model to generate a `model-friendly' calibration set, ensuring better alignment and mitigating quantization errors.}
\label{rob}
\end{figure}


Activation-distribution drift and quantization noise jointly degrade PTQ performance. Distribution drift arises as inputs vary across layers and during inference, causing mismatches between the quantization range (scale/zero-point) and the actual activations; this leads to accuracy loss. Moreover, quantization errors can propagate through nonlinearities, skip connections, and residuals, amplifying downstream degradation. Prior work mitigates these issues from multiple angles, including error-aware weight reconstruction (e.g., GPTQ~\cite{frantar2023gptq}), activation smoothing or scaling to tame outliers and range mismatch (e.g., SmoothQuant~\cite{xiao2024smoothquantaccurateefficientposttraining}, AWQ~\cite{lin2024awq}), and lightweight calibration-time refinement/correction with a few iterations to suppress accumulated errors (e.g., OmniQuant~\cite{shao2024omniquantomnidirectionallycalibratedquantization}, FPTQuant~\cite{breugel2025fptquant}).




However, many approaches fixate on the model level while neglecting calibration-sample adaptability. Calibration data may be constrained by privacy, and sample selection often requires multi-step optimization, introducing uncertainty. Real data closely tied to the target model may still fail to capture the complex internal activations during inference—especially in LLMs with prevalent outliers—leading to biased quantization parameters and notable accuracy degradation.

To address these challenges, we introduce \textbf{FAQ}, a \textbf{Q}uantization framework that optimizes calibration data by exploiting \textbf{F}amily \textbf{A}wareness. Models from the same development lineage and trained under similar paradigms exhibit substantial activation agreement; by reusing calibration mappings from “senior” family members, original samples with data bias can approximate the target model’s activations, achieving distribution alignment under blind conditions. The resulting calibrated samples are refined via inter-group competition and content normalization to form a high-quality calibration set, thereby mitigating PTQ-induced errors.


Our contributions are as follows:

\begin{itemize}
\item We propose FAQ, a framework that leverages intra-family activation consistency to optimize calibration samples. To our knowledge, this is the first work to exploit family priors in PTQ calibration data.
\item We show that family priors can be more influential for quantization performance than broader architectural similarities.
\item Our pipeline uses chain-of-thought (CoT) guidance and intra-group competition to constrain calibration-data optimization, enabling seamless integration of FAQ with traditional PTQ methods.
\item Extensive empirical comparisons with state-of-the-art (SOTA) PTQ baselines show that FAQ reduces quantization-induced accuracy loss by up to 28.5\%, evidencing superior performance.
\end{itemize}


\section{Related Work}

\subsection{Post-Training Quantization (PTQ)} 

Post-Training Quantization (PTQ) offers a low-cost, training-free solution for model compression but suffers significant accuracy degradation at low bit-widths, especially for LLMs~\cite{choukroun2019low,hubara2020improvingposttrainingneural,li2021brecqpushinglimitposttraining}. Research to address this challenge has largely followed two primary directions: innovations in the quantization algorithms and data-driven optimizations of the calibration set.

\textbf{Algorithm-level innovations} have predominantly focused on mitigating quantization errors. One major line of work involves mixed-precision schemes, where sensitive outlier weights are stored at higher precision while non-critical values are aggressively quantized~\cite{dettmers2024spqr, lee2024owq, ou2024amlq, kim2024squeezellm}. Another strategy is \textbf{activation-aware quantization}, which identifies and protects weights that are multiplied by large activation values, as they are more critical to model performance~\cite{lin2024awq, huang2025slimllm}. Others have focused on designing \textbf{advanced rounding mechanisms} or novel quantizers, such as additive quantization, to move beyond simple round-to-nearest schemes~\cite{lee2023flexround, chee2023quip, egiazarian2024extreme}. While powerful, these methods primarily treat the model's activation distribution as a fixed target to adapt to, rather than as a variable that can be optimized.

\textbf{Data-driven approaches} recognize the pivotal role of the calibration set. Early methods leveraged this data to perform layer-wise error minimization or block reconstruction~\cite{li2021brecq}. More recently, some works have focused on synthesizing calibration data from scratch when real data is unavailable, for instance, by matching batch normalization statistics~\cite{cai2020zeroq}. Our work, FAQ, advances this data-centric view. Instead of synthesizing data from statistical priors or merely using existing data to minimize local errors, we propose to regenerate a higher-quality calibration set from a more capable, in-family model, directly targeting the optimization of the activation distributions themselves.

\subsection{Quantization-Aware Training (QAT)}


Our approach should be distinguished from Quantization-Aware Training (QAT), which simulates quantization during a full fine-tuning phase~\cite{jacob2018quantization, esser2020learnedstepsizequantization}. While QAT can achieve superior accuracy, its high computational and data requirements are often prohibitive for LLMs. Our work remains strictly within the efficient, training-free PTQ paradigm, aiming to close the performance gap with QAT through data-centric innovations alone.

\begin{figure*}[t]
\centering
\includegraphics[width=2\columnwidth]{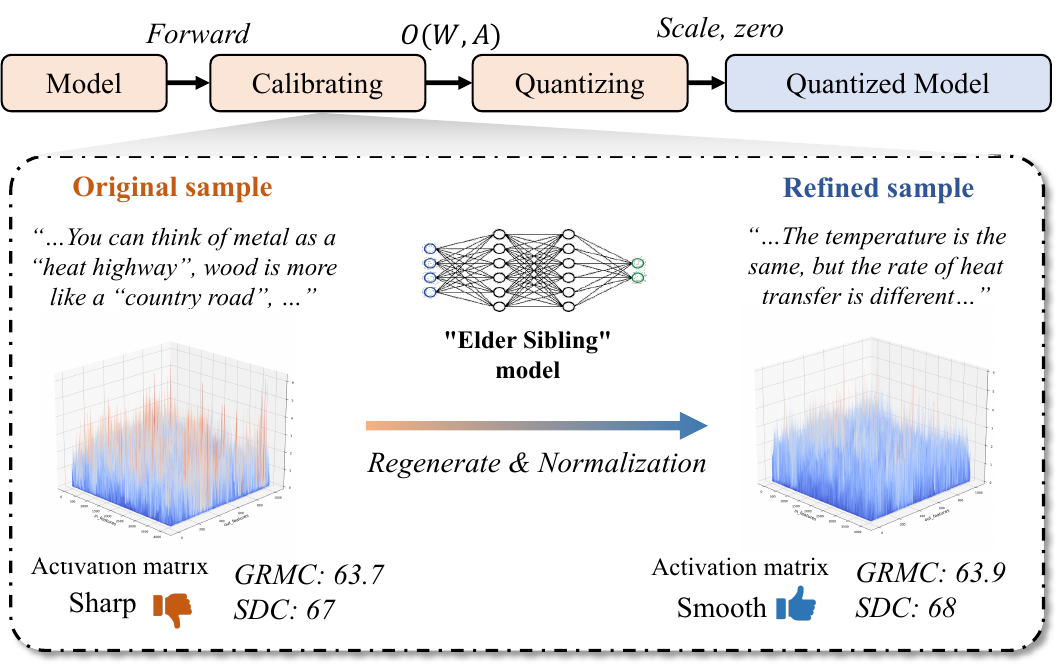}
\caption{FAQ-enhanced PTQ calibration.
Top: standard PTQ workflow. Bottom: zoom-in of the calibration stage. FAQ queries a larger in-family teacher (“elder-sibling”) model (Qwen3-235B-A22B) to regenerate and normalize the original calibration prompts, yielding refined calibration data. The refined set induces smoother activation statistics in the target model (Qwen3-8B), illustrated by fewer extreme peaks in the activations, and improves PTQ results on both GRMC and SDC. GRMC (General Reasoning and Multilingual Capabilities) is the average score over 12 general downstream tasks; SDC (Specialized Domain Capabilities: Math and Code) is the average accuracy over AIME, MATH-500, and LiveCodeBench.}
\label{3d-master}
\end{figure*}

 \section{Theoretical Framework}

\subsection{Motivation}
The fundamental goal of PTQ calibration is not to evaluate downstream task performance, which would require ground-truth labels, but rather to collect a small yet representative set of activations for the accurate computation of quantization parameters (e.g., scales and zero-points). From this standpoint, the critical factor is not the semantic fidelity of the calibration data to ground-truth, but the extent to which it can emulate the activation distributions the model encounters during inference.

Inspired by recent advanced methods such as GPTAQ~\cite{li2025gptaq}, we have found a key limitation of traditional quantizers like GPTQ~\cite{frantar2023gptq}: their symmetric calibration objective fails to account for the fact that the input activations $\mathbf{A}$ are themselves the quantized outputs of the preceding layers. Specifically, the quantization objective can be formulated as:
\begin{equation}
O(W, \mathbf{A}) = \min_{Q(\cdot) \in \mathcal{Q}} \left\lVert Q(W)\mathbf{A} - W\mathbf{A} \right\rVert_F^2 ,
\label{eq:gptq}
\end{equation}
where $W$ denotes the weight matrix, $\mathbf{A}$ represents the input activations, $Q(\cdot)$ is a quantization operator that maps $W$ to a lower-precision space, and $\mathcal{Q}$ denotes the set of all feasible quantized weights. However, in practice, $\mathbf{A}$ is not the original, full-precision activation, but rather the quantized output from the preceding layer, introducing a mismatch between the calibration objective and the actual inference process. This discrepancy, where the calibration input differs from the true full-precision activations $\tilde{\mathbf{A}}$, becomes more pronounced in deeper layers.
Therefore, a prevalent asymmetric calibration framework~\cite{frantar2023gptq} at this stage aims to correct for the propagated quantization error, which is formulated as:
\begin{equation}
O(W, \mathbf{A}, \tilde{\mathbf{A}}) = \min_{Q(\cdot) \in \mathcal{Q}} \|Q(W)\mathbf{A} - W \tilde{\mathbf{A}}\|_F^2 .
\label{eq:gptaq}
\end{equation}

While these algorithmic-level corrections are effective at managing challenging activation patterns, they are fundamentally compensatory in nature. The underlying issue is that the initial, full-precision activations $\tilde{\mathbf{A}}$ derived from standard calibration data are often inherently "hostile" to quantization, characterized by sparse, high-magnitude outliers. 
Our approach, FAQ, operates on a different and more fundamental premise. Instead of designing a more complex optimization objective to compensate for challenging activation patterns, we aim to proactively reshape the activation patterns themselves to be inherently more quantization-friendly. 

\subsection{Alignment of calibration data}

Compared to directly modifying the quantization algorithm, FAQ induces a superior calibration set $\mathcal{D}_{\text{FAQ}}$ to elicit a new set of full-precision activations, denoted as $\widehat{\mathbf{A}}$.

\begin{equation}
O(W_n, \mathbf{A}_n, \widehat{\mathbf{A}}) = \min_{Q(\cdot) \in \mathcal{Q}} \left\lVert Q(W_n)\mathbf{A}_n - W_n\mathbf{A}_n \right\rVert_F^2
\label{eq:faq_1}
\end{equation}

\begin{equation}
O(W_n, \mathbf{A}_n, \widehat{\mathbf{A}}_n) = \min_{Q(\cdot) \in \mathcal{Q}} \left\lVert Q(W_n)\mathbf{A}_n - W_n \widehat{\mathbf{A}}_n \right\rVert_F^2
\label{eq:faq_2}
\end{equation}

\begin{equation}
\mathbf{A}_{n+1} =  Q(W_n)\mathbf{A}_n, ~ \widehat{\mathbf{A}}_{n+1} =  W_n\widehat{\mathbf{A}}_n, ~ \mathbf{A}_0 = \widehat{\mathbf{A}}_0
\label{eq:faq_3}
\end{equation}

The effectiveness of FAQ stems from its ability to generate data that elicits more quantization-friendly activation patterns. When processing data $\widehat{\mathbf{A}}$ generated by FAQ, the activation distribution of a given layer is significantly smoother and more concentrated than the spiky and sparse distribution induced by the original calibration data $\tilde{\mathbf{A}}$, as visualized in Figure~\ref{3d-master}. By taming outliers at the data-source level, the quantization objective $O(W_n, \mathbf{A}_n, \widehat{\mathbf{A}}_n)$ becomes fundamentally easier to optimize. Consequently, even a standard, symmetric quantizer applied to our data can achieve superior performance, as described in Equation~(\ref{eq:faq_1}), Equation~(\ref{eq:faq_2}) and Equation~(\ref{eq:faq_3}). This is because the need for complex, asymmetric error correction is greatly diminished from the outset, making the subsequent quantization process more robust.

\section{Main Method}

We present FAQ, illustrated in Figure~\ref{rob}. The central idea of FAQ is to mitigate distributional mismatch between generic calibration data and a target model’s activation patterns. To achieve this, we replace the standard, pre-existing calibration set with a high-quality synthetic dataset specifically aligned with the target model's intrinsic characteristics.


\subsection{Calibration Regeneration}

We believe that models drawn from the same developmental lineage and trained under similar paradigms tend to exhibit consistent internal activations. This motivates our core hypothesis: calibration data regenerated by a more capable, in-family model are more effective for PTQ than real-world data alone. An in-family “Elder Sibling” model, sharing architecture and training paradigms with the target, implicitly encodes the target’s activation dynamics  (see Appendix Fig.~\ref{fig:three_acts}). Consequently, data it generates are not only semantically rich but also tailored to drive the target model's inference in a representative way, including surface-level outliers. This alignment yields more accurate quantization parameters and mitigates accuracy loss.


To exploit this prior, we introduce \textbf{In-family Synthesis with CoT}. For each query in the original seed calibration set, we prompt the larger, in-family "Elder Sibling" model to generate a new, detailed response. Crucially, we instruct the model to produce not just a final answer but also its intermediate reasoning process (e.g., via CoT prompting). This yields data with greater complexity and semantic diversity, designed to engage a broader set of pathways in the target model.


\subsection{Calibration Normalization}

The regeneration process produces a large number of synthetic, activation-distribution-aligned calibration samples. However, normalization is still necessary in practical applications to avoid the negative impact of unreasonable and unstable data on PTQ. 

First, we designed \textbf{Quality-driven Selection.} We generate three candidate responses per query and use a powerful external LLM (e.g., Qwen2.5-72B-Instruct~\cite{qwen2.5technicalreport}) as a judge to select the highest-scoring one, filtering out nonsensical generations. Secondly, \textbf{Template-based Data Assembly.} The selected response is combined with the original query and formatted using the target model's official chat template. This ensures the final sample's format perfectly aligns with the model's expected input, enhancing distributional consistency.

Together, regeneration and normalization convert a small seed set into a diverse, distributionally aligned calibration corpus that is structurally compatible with the target model, thereby enhancing PTQ performance.




\begin{figure}[t]
\centering
\includegraphics[width=1\columnwidth]{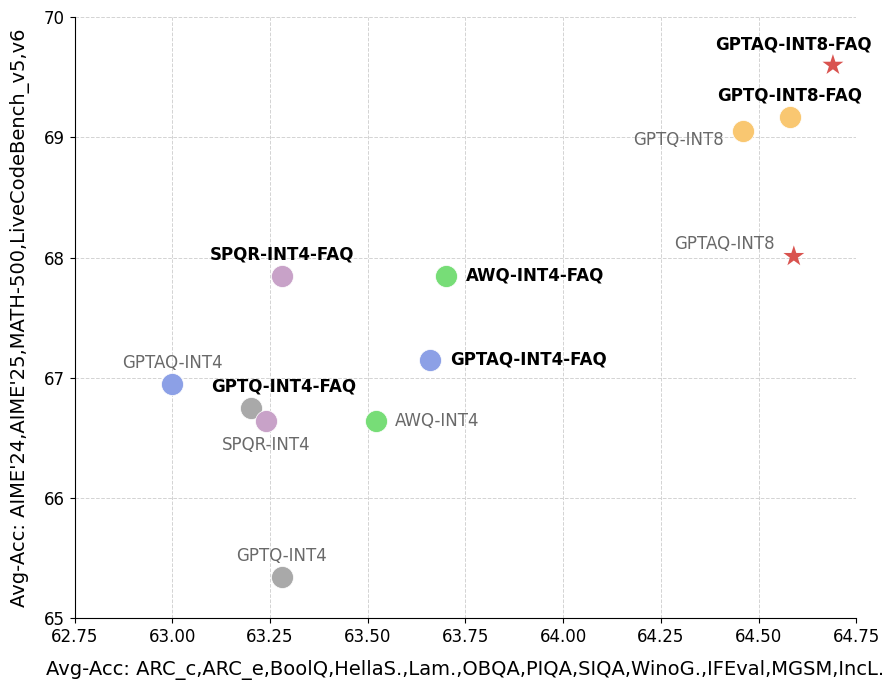} 

\caption{Overall performance improvement of FAQ across multiple quantization methods and benchmark suites on the Qwen3-8B model. X: average accuracy on general tasks; Y: performance on math/coding tasks. Each point is a quantization method (color = base algorithm, e.g., green for AWQ-INT4; gray vs bold label = baseline vs FAQ). The consistent up-right shift shows FAQ is a plug-and-play enhancement.}
    
\label{overall_performance}
\end{figure}

\section{Experiments}

In this section, we conduct a comprehensive set of experiments to validate FAQ's effectiveness and versatility as a plug-and-play enhancement for PTQ methods. To provide a holistic overview of our findings, we begin with Figure~\ref{overall_performance}, which encapsulates the main results on the Qwen3-8B model. As the plot demonstrates, applying our FAQ framework shifts the performance of baseline methods towards the top-right corner, signifying improved accuracy of the model. This overarching improvement provides the context for the detailed numerical results and analyses presented in the subsequent sections.

\subsection{Experimental Setup}

\textbf{Models and Methods.} We evaluate our framework on a diverse suite of recent open-source models from the Qwen3 family, including the dense Qwen3-8B, a reasoning-distilled variant, and the larger MoE-based Qwen3-30B-A3B. This allows us to assess performance across standard, specialized, and sparse architectures. We apply FAQ as a plug-and-play enhancement to four PTQ methods: GPTQ, AWQ, SPQR, and GPTAQ, primarily in INT4 and INT8 settings.

\textbf{Evaluation.} Our comprehensive evaluation covers three key areas:
\begin{enumerate}
    \item Language Modeling, measured by perplexity on Wikitext2~\cite{merity2016pointerwiki}, C4~\cite{raffel2020exploringc4}, and LAMBADA~\cite{radford2019language} benchmarks;
    \item General Reasoning and Multilingual Capabilities, assessed via the average accuracy on a broad suite of 12 downstream tasks:ARC-c and ARC-e~\cite{clark2018arc}, BoolQ~\cite{clark2019boolq}, Hellaswag~\cite{zellers2019hellaswag}, LAMBADA, OpenBookQA~\cite{mihaylov2018obqa}, PIQA~\cite{bisk2020piqa}, SIQA~\cite{sap2019socialiqa}, Winogrande~\cite{sakaguchi2021winogrande}, IFEval~\cite{zhou2023ifeval}, MGSM~\cite{shi2022mgsm} and INCLUDE~\cite{romanou2024include};
    \item Specialized Domain Capabilities, tested on challenging math and code generation benchmarks: AIME'24 and AIME'25~\cite{aime}, MATH-500~\cite{lightman2023math500}, and LiveCodeBench~\cite{jain2024livecodebench}.
\end{enumerate}
All downstream evaluations are conducted in a zero-shot setting. To ensure robustness, all reported scores are averaged over multiple runs. A detailed description of models, PTQ configurations, benchmark lists, and our reliability assurance protocol can be found in Appendix \ref{appendix:detailed_experimental_setup}.

\begin{table}[h]
    \centering
            {
    \small          

    \renewcommand{\arraystretch}{1.1}
    \begin{tabular}{@{}l l c c c }
        \toprule
        \multirow{2}{*}{\textbf{Bits}} & \multirow{2}{*}{\textbf{Method}} & \multicolumn{3}{c}{\textbf{Qwen3-8B}} \\
        \cmidrule(lr){3-5}
         &  & \textbf{Wiki2}(\textdownarrow) & \textbf{C4}(\textdownarrow) & \textbf{Lambada}(\textdownarrow) \\
        \midrule
        BF16 & - & 12.20 & 36.37 & 6.19 \\
        \midrule
        \multirow{4}{*}{INT8} & GPTQ & 12.20 & 36.36 & 6.25 \\
         & \cellcolor{gray!20}+FAQ & \cellcolor{gray!20}\textbf{12.19} & \cellcolor{gray!20}\textbf{36.33} & \cellcolor{gray!20}\textbf{6.23} \\ 
         & GPTAQ & 12.22 & \textbf{36.37} & 6.21 \\ 
         & \cellcolor{gray!20}+FAQ & \cellcolor{gray!20}\textbf{12.22} & \cellcolor{gray!20}36.40 & \cellcolor{gray!20}\textbf{6.16} \\ 
         \midrule
        \multirow{8}{*}{INT4} & GPTQ & 13.10 & 38.97 & 7.55 \\
         & \cellcolor{gray!20}+FAQ & \cellcolor{gray!20}\textbf{12.95} & \cellcolor{gray!20}\textbf{38.62} & \cellcolor{gray!20}\textbf{7.26} \\ 
         & AWQ & 12.80 & 38.72 & \textbf{6.90} \\ 
         & \cellcolor{gray!20}+FAQ & \cellcolor{gray!20}\textbf{12.73} & \cellcolor{gray!20}\textbf{38.60} & \cellcolor{gray!20}7.16 \\ 
         & SPQR & 16.40 & 56.70 & \textbf{7.37} \\
         & \cellcolor{gray!20}+FAQ & \cellcolor{gray!20}\textbf{14.95} & \cellcolor{gray!20}\textbf{46.27} & \cellcolor{gray!20}8.03 \\ 
         & GPTAQ & 13.01 & 39.17 & 6.88 \\
         & \cellcolor{gray!20}+FAQ & \cellcolor{gray!20}\textbf{12.99} & \cellcolor{gray!20}\textbf{39.01} & \cellcolor{gray!20}\textbf{6.62} \\ 
         \midrule
         \multirow{2}{*}{AVG} & Quant & 13.29 & 41.05 & \textbf{6.86} \\ 
         & \cellcolor{blue!10}+FAQ & \cellcolor{blue!10}\textbf{13.01} & \cellcolor{blue!10}\textbf{39.21} & \cellcolor{blue!10}6.91 \\ 
        \bottomrule
    \end{tabular}
    }
    \caption{Perplexity comparison of different quantization methods on Qwen3-8B models. The lower is the better.}
    \label{tab:qwen3_8b_perplexity}
\end{table}

\begin{table*}[h]
    \centering
            {
    \small          

    \renewcommand{\arraystretch}{1.2}
    \begin{tabular}{@{}>{\centering\arraybackslash}p{0.03\linewidth}>{\centering\arraybackslash}p{0.05\linewidth}>{\centering\arraybackslash}p{0.04\linewidth}>{\centering\arraybackslash}p{0.04\linewidth}>{\centering\arraybackslash}p{0.04\linewidth}>{\centering\arraybackslash}p{0.05\linewidth}>{\centering\arraybackslash}p{0.035\linewidth}>{\centering\arraybackslash}p{0.05\linewidth}>{\centering\arraybackslash}p{0.035\linewidth}>{\centering\arraybackslash}p{0.035\linewidth}>{\centering\arraybackslash}p{0.05\linewidth}>{\centering\arraybackslash}p{0.045\linewidth}>{\centering\arraybackslash}p{0.045\linewidth}>{\centering\arraybackslash}p{0.04\linewidth}>{\centering\arraybackslash}p{0.035\linewidth}}
        \toprule
        \textbf{Bits} & \textbf{Method} & \textbf{ARC\_c} & \textbf{ARC\_e} & \textbf{BoolQ}& \textbf{HellaS.}& \textbf{Lam.}& \textbf{OBQA}& \textbf{PIQA}& \textbf{SIQA}& \textbf{WinoG.}& \textbf{IFEval}& \textbf{MGSM} & \textbf{IncL.}&\textbf{Avg}\\
        \midrule
         BF16& -& 55.2& 83.6& 86.7& 57.1& 61.6& 31.4& 76.6& 41.8& 68.0&84.8& 54.6&75.4&64.7 \\
         \midrule
         \multirow{4}{*}{INT8}& GPTQ & \textbf{55.1}& \textbf{83.5}& 86.6& \textbf{57.2}& 61.4& \textbf{31.4}& 76.8& 41.5& 67.7&83.0& 54.0&75.4& 64.5 \\ 
         & \cellcolor{gray!20}+FAQ& \cellcolor{gray!20}55.0& \cellcolor{gray!20}83.4& \cellcolor{gray!20}\textbf{86.6}& \cellcolor{gray!20}57.1& \cellcolor{gray!20}\textbf{61.5}& \cellcolor{gray!20}31.1& \cellcolor{gray!20}\textbf{76.9}& \cellcolor{gray!20}\textbf{41.6}& \cellcolor{gray!20}\textbf{68.1}&\cellcolor{gray!20}\textbf{83.2}& \cellcolor{gray!20}\textbf{54.9}&\cellcolor{gray!20}\textbf{75.8}&\cellcolor{gray!20}\textbf{64.6} \\
         & GPTAQ & 55.0& 83.6& \textbf{86.5}& \textbf{57.1}& 61.4& \textbf{31.6}& 76.6& 41.4&\textbf{68.3}& 83.3& 54.5&\textbf{75.8}& 64.6 \\ 
         & \cellcolor{gray!20}+FAQ& \cellcolor{gray!20}\textbf{55.5}& \cellcolor{gray!20}\textbf{83.7}& \cellcolor{gray!20}86.4& \cellcolor{gray!20}\textbf{57.1}& \cellcolor{gray!20}\textbf{61.7}& \cellcolor{gray!20}31.5& \cellcolor{gray!20}\textbf{76.7}& \cellcolor{gray!20}\textbf{41.6}& \cellcolor{gray!20}68.0&\cellcolor{gray!20}\textbf{83.8}&\cellcolor{gray!20}\textbf{54.7}&\cellcolor{gray!20}75.6&\cellcolor{gray!20}\textbf{64.7} \\
        \midrule
         \multirow{8}{*}{INT4}& GPTQ & \textbf{52.7}& 80.9& \textbf{86.3}& 55.9& 59.0& 29.8& 76.3& 41.1&\textbf{68.5}& 81.3 &\textbf{56.4}&\textbf{71.2}&\textbf{63.3} \\
         & \cellcolor{gray!20}+FAQ& \cellcolor{gray!20}51.8& \cellcolor{gray!20}\textbf{81.6}& \cellcolor{gray!20}85.8& \cellcolor{gray!20}\textbf{55.9}& \cellcolor{gray!20}\textbf{59.6}& \cellcolor{gray!20}\textbf{31.2}& \cellcolor{gray!20}\textbf{76.8}& \cellcolor{gray!20}\textbf{42.1}& \cellcolor{gray!20}68.2&\cellcolor{gray!20}\textbf{82.1}&\cellcolor{gray!20}53.7&\cellcolor{gray!20}69.7&\cellcolor{gray!20}63.2 \\
         & AWQ & 52.8& 81.4&\textbf{86.7}& 56.0& 60.1& 30.2&\textbf{76.2}&\textbf{41.0}&\textbf{67.5}& 82.1& 54.2&\textbf{73.9}& 63.5 \\
         & \cellcolor{gray!20}+FAQ& \cellcolor{gray!20}\textbf{53.1}&\cellcolor{gray!20}\textbf{81.6}&\cellcolor{gray!20}86.4&\cellcolor{gray!20}\textbf{56.0}&\cellcolor{gray!20}\textbf{60.6}&\cellcolor{gray!20}\textbf{30.8}&\cellcolor{gray!20}76.1&\cellcolor{gray!20}40.5&\cellcolor{gray!20}67.0&\cellcolor{gray!20}\textbf{83.6}&\cellcolor{gray!20}\textbf{56.6}&\cellcolor{gray!20}71.8&\cellcolor{gray!20}\textbf{63.7} \\
         & SPQR & \textbf{54.5}&\textbf{83.0}& 85.9 &\textbf{56.4}& 59.3& 29.8 & 76.0 &\textbf{40.8}&66.8&82.8&\textbf{53.3}&70.3&63.2 \\ 
         & \cellcolor{gray!20}+FAQ&\cellcolor{gray!20}53.5&\cellcolor{gray!20}80.9&\cellcolor{gray!20}\textbf{86.1}&\cellcolor{gray!20}56.0&\cellcolor{gray!20}\textbf{59.8}&\cellcolor{gray!20}\textbf{30.4}&\cellcolor{gray!20}\textbf{77.3}&\cellcolor{gray!20}40.5&\cellcolor{gray!20}\textbf{68.4}&\cellcolor{gray!20}\textbf{83.6}&\cellcolor{gray!20}51.8&\cellcolor{gray!20}\textbf{71.1}&\cellcolor{gray!20}\textbf{63.3} \\
         & GPTAQ & 53.0& 81.5 & 85.3& 55.3&\textbf{60.1}&\textbf{31.6}& 75.7& 41.0 & 66.9& 82.3 & 51.4 & 71.9& 63.0 \\ 
        & \cellcolor{gray!20}+FAQ&\cellcolor{gray!20}\textbf{53.8}&\cellcolor{gray!20}\textbf{82.0}&\cellcolor{gray!20}\textbf{86.4}&\cellcolor{gray!20}\textbf{55.8}&\cellcolor{gray!20}60.0&\cellcolor{gray!20}29.6&\cellcolor{gray!20}\textbf{75.9}&\cellcolor{gray!20}\textbf{41.0}&\cellcolor{gray!20}\textbf{68.6}&\cellcolor{gray!20}\textbf{83.1}&\cellcolor{gray!20}\textbf{53.8}&\cellcolor{gray!20}\textbf{74.0}&\cellcolor{gray!20}\textbf{63.7} \\
         \midrule
          \multirow{2}{*}{AVG}& Quant & \textbf{53.9}&\textbf{82.3}& 86.2& 56.3 & 60.2 & 30.7 & 76.2 & 41.1 & 67.6 & 82.5 & 54.0 &\textbf{73.1}& 63.7 \\ 
         & \cellcolor{blue!10}+FAQ&\cellcolor{blue!10}53.8&\cellcolor{blue!10}82.2&\cellcolor{blue!10}\textbf{86.3}&\cellcolor{blue!10}\textbf{56.3}&\cellcolor{blue!10}\textbf{60.5}&\cellcolor{blue!10}\textbf{30.8}&\cellcolor{blue!10}\textbf{76.6}&\cellcolor{blue!10}\textbf{41.2}&\cellcolor{blue!10}\textbf{68.1}&\cellcolor{blue!10}\textbf{83.2}&\cellcolor{blue!10}\textbf{54.2}&\cellcolor{blue!10}73.0&\cellcolor{blue!10}\textbf{63.9} \\
        \bottomrule
    \end{tabular}
    }
    \caption{Performance on 12 general downstream tasks with different quantization methods on Qwen3-8B models. Higher is better. The positive trend in the average score suggests that the benefits of FAQ are systematic.}
    \label{tab:qwen3_8b_fst_bolded_corrected}
\end{table*}

\subsection{Main Results}

\subsubsection{Language Modeling Performance.}
We first evaluate the impact of FAQ on the fundamental language modeling capabilities of the quantized model, measured by Perplexity (PPL) on the Wikitext2, C4, and LAMBADA datasets. Lower PPL indicates better performance. The detailed results for the Qwen3-8B model are presented in Table~\ref{tab:qwen3_8b_perplexity}.
The results demonstrate that FAQ enhances the language modeling performance of all evaluated PTQ methods across both INT8 and INT4 quantization in most cases. The average (AVG) PPL across all three benchmarks is consistently reduced with FAQ. For instance, in the more challenging INT4 setting, FAQ brings a substantial PPL reduction for SPQR on the C4 dataset, decreasing it from 56.70 to 46.27. Similarly, for GPTQ-INT4 and GPTAQ-INT4, FAQ improves performance across all three datasets, highlighting its robust and broad effectiveness.
This consistent improvement can be attributed to the higher-quality calibration data generated by FAQ. By producing a calibration set that better reflects the model's typical activation distributions and tames outlier features, FAQ enables the quantizer to learn more accurate scaling factors. This leads to lower quantization error and, consequently, a better preservation of the model's nuanced understanding of language, as evidenced by the lower perplexity scores.

\subsubsection{General Reasoning and Multilingual Capabilities.}

To assess whether improvements in language modeling translate to broader cognitive abilities, we evaluate the models on a diverse suite of 12 downstream tasks. As detailed in Table~\ref{tab:qwen3_8b_fst_bolded_corrected}, the results demonstrate the wide-ranging benefits of FAQ. On average, applying FAQ leads to performance gains, reducing the accuracy loss compared to the full-precision baseline by 28.5\%. This positive trend is particularly notable on reasoning-intensive benchmarks like IFEval (instruction-following) and MGSM (multilingual math), where FAQ helps recover significant performance otherwise lost to quantization. This suggests that FAQ's enhanced calibration preserves not just language modeling fidelity but also the critical model parameters responsible for complex reasoning, enabling the quantized model to better retain its general-purpose problem-solving skills.

We note a marginal 0.1-point average accuracy dip for the GPTQ-INT4 configuration, which we trace to a significant performance drop solely on the MGSM benchmark. We hypothesize this is a corner-case interaction where our regeneration process, in an otherwise beneficial act of smoothing distributions for the highly aggressive GPTQ-INT4 quantizer, may have inadvertently altered outlier features that were coincidentally crucial for MGSM's specific numerical reasoning patterns. This finding highlights a complex trade-off in extreme quantization scenarios, but the strong positive trend across all other methods and benchmarks confirms the substantial net benefit of FAQ.

\subsubsection{Specialized Domain Capabilities: Math and Code.}

Finally, to rigorously probe the limits of quantized models, we evaluate them on highly challenging math (AIME, MATH-500) and code generation (LiveCodeBench) benchmarks, which are notoriously sensitive to precision loss. The results in Table~\ref{tab:qwen3_8b_quantization_aime_math_lcb} demonstrate the remarkable resilience FAQ provides. Quantitative results show FAQ improves accuracy from 67.0 to 68.0 while reducing quantization-induced accuracy loss by 22\%, conclusively demonstrating that our data regeneration strategy effectively preserves models' complex reasoning abilities.

The benefits are particularly evident in the most aggressive INT4 setting. For instance, FAQ boosts the average accuracy of AWQ by 1.3 points (66.6 to 67.9) and SPQR by a notable 1.4 points, with consistent gains observed across individual benchmarks like AIME and LiveCodeBench.
This strong performance provides compelling evidence for our core hypothesis: complex reasoning relies on high-magnitude activations that represent critical operations. By generating data that tames the distribution of these crucial features, FAQ prevents their distortion during quantization, ensuring the essential building blocks of the model's reasoning capabilities are kept intact, even under extreme compression.

\begin{table}[h]
    \centering
        {
    \small          

    \renewcommand{\arraystretch}{1.3} 
    \begin{tabular}{@{}>{\centering\arraybackslash}p{0.075\linewidth}>{\centering\arraybackslash}p{0.1\linewidth} >{\centering\arraybackslash}p{0.065\linewidth} >{\centering\arraybackslash}p{0.065\linewidth} >{\centering\arraybackslash}p{0.08\linewidth} >{\centering\arraybackslash}p{0.06\linewidth} >{\centering\arraybackslash}p{0.06\linewidth} >{\centering\arraybackslash}p{0.07\linewidth}}
        \toprule
         \multirow{2}{*}{\textbf{Bits}} & \multirow{2}{*}{\textbf{Method}} & \multicolumn{2}{c}{\textbf{AIME}}& \textbf{Math}& \multicolumn{2}{c}{\textbf{L.C.B.}}& \multirow{2}{*}{\textbf{Avg}}\\
          \cmidrule(lr){3-4} \cmidrule(lr){5-5} \cmidrule(lr){6-7}& &\textbf{24}(\textuparrow)& \textbf{25}(\textuparrow)& \textbf{500}(\textuparrow)& \textbf{v5}(\textuparrow)& \textbf{v6}(\textuparrow)& \\
        \midrule 
         BF16 & - & 83.3& 76.7& 95.2& 58.1& 52.6& 73.2
\\
        \midrule 
         \multirow{4}{*}{INT8} 
         & GPTQ & 76.3& \textbf{68.3}& 94.6& 57.2& 48.9&69.1
\\
         & \cellcolor{gray!15} +FAQ & \cellcolor{gray!15} \textbf{79.2} & \cellcolor{gray!15} 62.9 & \cellcolor{gray!15} \textbf{95.1} & \cellcolor{gray!15} \textbf{58.7} & \cellcolor{gray!15} \textbf{50.0} & \cellcolor{gray!15} \textbf{69.2} 
\\
           & GPTAQ & 74.6& 68.8& \textbf{94.7}& 53.3& 48.7& 68.0
\\
          & \cellcolor{gray!15}+FAQ & \cellcolor{gray!15}\textbf{78.3}& \cellcolor{gray!15}\textbf{69.2}& \cellcolor{gray!15}94.2& \cellcolor{gray!15}\textbf{56.3}& \cellcolor{gray!15}\textbf{50.0}& \cellcolor{gray!15}\textbf{69.6}
\\
        \midrule 
         \multirow{8}{*}{INT4} 
         & GPTQ & \textbf{74.2}& 59.2& 95.0& \textbf{52.1}& 46.3& 65.3
\\
         & \cellcolor{gray!15}+FAQ & \cellcolor{gray!15}73.3& \cellcolor{gray!15}\textbf{66.7}& \cellcolor{gray!15}\textbf{95.1}& \cellcolor{gray!15}51.5& \cellcolor{gray!15}\textbf{47.2}& \cellcolor{gray!15}\textbf{66.8}
\\
           & AWQ & 72.5& 63.3& 94.9& 53.9& 48.6& 66.6
\\
          & \cellcolor{gray!15}+FAQ & \cellcolor{gray!15}\textbf{75.4}& \cellcolor{gray!15}\textbf{65.0}& \cellcolor{gray!15}\textbf{94.9}& \cellcolor{gray!15}\textbf{55.1}& \cellcolor{gray!15}\textbf{48.9}& \cellcolor{gray!15}\textbf{67.9}
\\
           & SPQR & 74.6& 65.6& 94.2& 50.6& 45.4& 66.1
\\
          & \cellcolor{gray!15}+FAQ & \cellcolor{gray!15}\textbf{76.7}& \cellcolor{gray!15}\textbf{65.6}& \cellcolor{gray!15}\textbf{94.4}& \cellcolor{gray!15}\textbf{54.2}& \cellcolor{gray!15}\textbf{46.4}& \cellcolor{gray!15}\textbf{67.5}
\\
           & GPTAQ & \textbf{78.3}& 62.5& 94.8& 52.7& 46.4& 67.0
\\
         & \cellcolor{gray!15}+FAQ & \cellcolor{gray!15}75.0& \cellcolor{gray!15}\textbf{64.4}& \cellcolor{gray!15}\textbf{94.8}& \cellcolor{gray!15}\textbf{53.3}& \cellcolor{gray!15}\textbf{48.2}& \cellcolor{gray!15}\textbf{67.2}
\\
         \midrule 
          \multirow{2}{*}{AVG} & Quant & 75.1& 64.6& 94.7& 53.3& 47.4& 67.0
\\
         & \cellcolor{blue!10}+FAQ & \cellcolor{blue!10}\textbf{76.3}& \cellcolor{blue!10}\textbf{65.6}& \cellcolor{blue!10}\textbf{94.8}& \cellcolor{blue!10}\textbf{54.8}& \cellcolor{blue!10}\textbf{48.4}& \cellcolor{blue!10}\textbf{68.0}\\
        \bottomrule
    \end{tabular}
    }
    \caption{Accuracy on specialized math and code benchmarks for the Qwen3-8B model. FAQ's consistent performance improvement, especially in the challenging INT4 setting, demonstrates its ability to preserve critical reasoning capabilities under extreme compression. Higher is better.}
    \label{tab:qwen3_8b_quantization_aime_math_lcb}
\end{table}

\subsection{Ablation Studies and Generalization}

Having established the significant and consistent benefits of FAQ on the Qwen3-8B model, we now turn to two critical questions to further understand its underlying principles and scope of applicability. First, we conduct an ablation study to validate our core "Family-Aware" hypothesis. Second, we test the generalization of FAQ to a larger-scale, more complex MoE model to assess its scalability.


\begin{table}[h]
    \centering
        {
    \small          
    \setlength{\tabcolsep}{5.5 pt}

    \renewcommand{\arraystretch}{1.3}
    \begin{tabular}{@{}>{\raggedright\arraybackslash}p{0.07\linewidth}>{\raggedright\arraybackslash}p{0.18\linewidth}>{\centering\arraybackslash}p{0.08\linewidth}>{\centering\arraybackslash}p{0.07\linewidth}>{\centering\arraybackslash}p{0.1\linewidth}|c>{\centering\arraybackslash}p{0.07\linewidth}}
        \toprule
        \multirow{2}{*}{\textbf{Bits}} & \multirow{2}{*}{\textbf{Method}} & \multicolumn{5}{c}{\textbf{DeepSeek-R1-0528-Qwen3-8B}}\\
        \cmidrule(lr){3-7}&  & \textbf{Wiki2} & \textbf{C4} & \textbf{Lam.}  & \textbf{GRMC}& \textbf{SDC}\\
        \midrule
        BF16 & - & 13.23& 40.83& 27.32& 60.64& 71.57
\\
        \midrule
        \multirow{4}{*}{INT8} & GPTQ+ds & 13.24& \textbf{40.85}& 27.42& 60.74& 70.44
\\
         & \cellcolor{gray!15}GPTQ+qw & \cellcolor{gray!15}\textbf{13.24}& \cellcolor{gray!15}40.86& \cellcolor{gray!15}\textbf{27.41}& \cellcolor{gray!15}\textbf{60.77}& \cellcolor{gray!15}\textbf{70.89}
\\ 
         & GPTAQ+ds& \textbf{13.20}& 40.73& 27.32& 60.42& 69.80
\\ 
         & \cellcolor{gray!15}GPTAQ+qw & \cellcolor{gray!15}13.21& \cellcolor{gray!15}\textbf{40.72}& \cellcolor{gray!15}\textbf{27.12}& \cellcolor{gray!15}\textbf{60.51}& \cellcolor{gray!15}\textbf{69.93}
\\ 
         \midrule
        \multirow{8}{*}{INT4} & GPTQ+ds& 14.30& 44.27& 33.95& 58.26& 67.14
\\
         & \cellcolor{gray!15}GPTQ+qw & \cellcolor{gray!15}\textbf{14.22}& \cellcolor{gray!15}\textbf{44.24}& \cellcolor{gray!15}\textbf{31.90}& \cellcolor{gray!15}\textbf{58.27}& \cellcolor{gray!15}\textbf{68.26}
\\ 
         & AWQ+ds& 13.93& 43.29& 33.65& 59.19& 66.83
\\ 
         & \cellcolor{gray!15}AWQ+qw & \cellcolor{gray!15}\textbf{13.87}& \cellcolor{gray!15}\textbf{43.00}& \cellcolor{gray!15}\textbf{30.11}& \cellcolor{gray!15}\textbf{59.27}& \cellcolor{gray!15}\textbf{68.20}
\\ 
         & SPQR+ds& 14.36& \textbf{44.18}& 34.61& 58.00& 65.16
\\
         & \cellcolor{gray!15}SPQR+qw & \cellcolor{gray!15}\textbf{14.35}& \cellcolor{gray!15}44.58& \cellcolor{gray!15}\textbf{29.77}& \cellcolor{gray!15}\textbf{58.28}& \cellcolor{gray!15}\textbf{67.11}
\\ 
         & GPTAQ+ds& 14.37& \textbf{44.12}& 33.66& 58.98& 67.69
\\
         & \cellcolor{gray!15}GPTAQ+qw & \cellcolor{gray!15}\textbf{14.31}& \cellcolor{gray!15}44.18& \cellcolor{gray!15}\textbf{29.89}& \cellcolor{gray!15}\textbf{59.65}& \cellcolor{gray!15}\textbf{68.03}
\\ 
         \midrule
         \multirow{2}{*}{AVG} & Quant+ds& 13.90& \textbf{42.91}& 31.77& 59.27& 67.84
\\ 
         & \cellcolor{blue!10}Quant+qw & \cellcolor{blue!10}\textbf{13.87}& \cellcolor{blue!10}42.93& \cellcolor{blue!10}\textbf{29.37}& \cellcolor{blue!10}\textbf{59.46}& \cellcolor{blue!10}\textbf{68.74}\\ 
        \bottomrule
    \end{tabular}
    } 

    \caption{Performance comparison of Family-Sourced (+qw) vs. Knowledge-Sourced (+ds) calibration data on the distilled DeepSeek-R1-0528-Qwen3-8B model. Lower perplexity is better; higher accuracy is better (GRMC/SDC). +qw consistently outperforms +ds, supporting the importance of within-family alignment for FAQ.}
    \label{tab:deepseek_r1_0528_qwen3_8b_eval}
\end{table}

\subsubsection{Validating the "Family-Aware" Hypothesis.}

\textbf{A Unique Testbed: The Distilled Model.} Our core hypothesis posits that the efficacy of FAQ stems from a deep "family" connection, which we define not by macro-architectural identity (e.g., MoE vs. Dense) but by a shared developmental lineage. This includes overlapping training corpora, consistent tokenization schemes, and a common design philosophy for fundamental building blocks. To rigorously test this, we leverage a unique model: DeepSeek-R1-0528-Qwen3-8B. This model was created by distilling knowledge from a "knowledge teacher," DeepSeek-R1 (an MoE model), into a "student model," Qwen3-8B (a Dense model).
This setup creates a fascinating and powerful adversarial test. We compare two data generation strategies, both using large MoE models as generators:
\begin{enumerate}
    \item Teacher-Sourced Calibration (+ds): Generating data using the knowledge teacher, DeepSeek-R1.
    \item Family-Sourced Calibration (+qw): Generating data using the student's family model, Qwen3-235B-A22B.
\end{enumerate}

\textbf{Results and Analysis.} The results in Table~\ref{tab:deepseek_r1_0528_qwen3_8b_eval} provide a decisive answer to the central question of whether knowledge origin (DeepSeek) or developmental lineage (Qwen) is more effective. The Family-Sourced (+qw) calibration decisively outperforms the Teacher-Sourced (+ds) data, improving the average score on specialized tasks (68.74 vs. 67.84) and reducing perplexity on Lambada (29.37 vs. 31.77).

This strong head-to-head comparison validates our core hypothesis: a shared "developmental lineage" is more critical than knowledge origin. The superior performance is not attributable to a naive architectural match—as both generators are MoE while the student is Dense—but rather stems from a form of "distributional homology". This homology enables the generator to craft inputs that elicit more natural, quantization-friendly activation patterns. Due to space constraints, we present the averaged results here; a detailed, per-benchmark comparison can be found in Appendix.

\begin{table}[h]
    \centering

    {
    \small          

    \setlength{\tabcolsep}{5.5 pt}

    \renewcommand{\arraystretch}{1.2}
    \begin{tabular}{@{}l l >{\centering\arraybackslash}p{0.09\linewidth}>{\centering\arraybackslash}p{0.07\linewidth}>{\centering\arraybackslash}p{0.1\linewidth}|cc}
        \toprule
        \multirow{2}{*}{\textbf{Bits}} & \multirow{2}{*}{\textbf{Method}} & \multicolumn{5}{c}{\textbf{Qwen3-30B-A3B}}\\
        \cmidrule(lr){3-7}&  & \textbf{Wiki2} & \textbf{C4} & \textbf{Lam.}  & \textbf{GRMC}& \textbf{SDC}\\
        \midrule
        BF16 & - & 10.89& 30.13& 5.44
& 66.85& 74.37
\\
        \midrule
        \multirow{2}{*}{INT8} & GPTQ & 11.02& 30.84& 5.58
& 66.55& 71.50
\\
         & \cellcolor{gray!15}+FAQ & \cellcolor{gray!15}\textbf{11.02}& \cellcolor{gray!15}\textbf{30.84}& \cellcolor{gray!15}\textbf{5.56}
& \cellcolor{gray!15}\textbf{66.81}& \cellcolor{gray!15}\textbf{72.29}
\\ 
         \midrule
        \multirow{4}{*}{INT4} & GPTQ & 11.36& 31.87& 6.21
& 65.37& 71.53
\\
         & \cellcolor{gray!15}+FAQ & \cellcolor{gray!15}\textbf{11.32}& \cellcolor{gray!15}\textbf{31.60}& \cellcolor{gray!15}\textbf{6.14}& \cellcolor{gray!15}\textbf{65.53}& \cellcolor{gray!15}\textbf{71.78}
\\ 
         & SPQR & 13.08& 40.49& 6.55& 65.10& 70.69
\\
         & \cellcolor{gray!15}+FAQ & \cellcolor{gray!15}\textbf{11.83}& \cellcolor{gray!15}\textbf{34.81}& \cellcolor{gray!15}\textbf{6.01}& \cellcolor{gray!15}\textbf{65.36}& \cellcolor{gray!15}\textbf{71.66}
\\ 
         \midrule
         \multirow{2}{*}{AVG} & Quant & 11.82& 34.40& 6.11
& 65.67& 71.24
\\ 
         & \cellcolor{blue!10}+FAQ & \cellcolor{blue!10}\textbf{11.39}& \cellcolor{blue!10}\textbf{32.42}& \cellcolor{blue!10}\textbf{5.90}& \cellcolor{blue!10}\textbf{65.90}& \cellcolor{blue!10}\textbf{71.91}\\ 
        \bottomrule
    \end{tabular}
            } 

    \caption{Generalization of FAQ to the larger-scale, MoE-based Qwen3-30B-A3B model. We compare baseline PTQ methods and their FAQ-enhanced variants; FAQ consistently improves perplexity and GRMC/SDC accuracy, demonstrating scalability to larger, more complex architectures.}
    \label{tab:qwen3_30b_eval}
\end{table}

\subsubsection{Generalization to MoE Architectures.}

To demonstrate FAQ's scalability and versatility, we apply it to the Qwen3-30B-A3B, a larger MoE model whose sparse, gated activations present a rigorous test. The results in Table~\ref{tab:qwen3_30b_eval} robustly mirror our earlier success. On average, FAQ boosts the specialized task score from 67.84 to 68.74, confirming its effectiveness on complex architectures.

The performance lift is particularly pronounced in the demanding INT4 setting, where FAQ boosts SPQR's accuracy by nearly two points (65.16 to 67.11) and GPTQ's from 67.14 to 68.26. This is also reflected in a substantial perplexity reduction on Lambada. This successful application to a 30B MoE model validates that FAQ is a robust and generalizable framework, whose benefits are not confined to model size or density but extend effectively to sparse activation environments, making it a broadly applicable tool for modern LLMs.

\section{Conclusions}
\label{sec:Conclusions}

We introduce FAQ, a \textbf{Q}uantization framework that optimizes calibration data by exploiting \textbf{F}amily \textbf{A}wareness. FAQ addresses a fundamental source of quantization error—outlier-prone activations—by leveraging an in-family, larger “Elder Sibling” model to regenerate a high-quality, quantization-friendly calibration set. Comprehensive experiments demonstrate that FAQ acts as a universal, plug-and-play performance booster across a broad spectrum of SOTA PTQ methods, model sizes, and architectures, including MoE models. Crucially, our ablation study provides direct evidence that FAQ’s effectiveness stems from a shared developmental lineage between models, a factor more critical than the origin of knowledge or macro-architectural similarity. Collectively, this work inaugurates a data-centric paradigm for PTQ and highlights the substantial potential of leveraging model family priors to yield more robust and efficient LLMs.

\section*{Limitations}
\label{sec:Conclusions}
The effectiveness of FAQ can be influenced by specific choices in the regeneration process, such as prompting strategies and generation hyperparameters (e.g., temperature, candidate selection). While our default settings demonstrate broad robustness, extreme quantization regimes (e.g., INT4) inherently introduce volatility. Consequently, minor performance fluctuations may occur in isolated edge cases—such as the marginal degradation observed on MGSM—suggesting a valuable avenue for future research into adaptive calibration strategies to further enhance stability.



\bibliography{custom}

\appendix

\section{Appendix}
\label{sec:appendix}


\begin{table*}[htbp!]
    \centering
        {
    \small          
    \setlength{\tabcolsep}{5 pt}
        
    \renewcommand{\arraystretch}{1.2}
    \begin{tabular}{@{}l>{\centering\arraybackslash}p{0.04\linewidth}c>{\centering\arraybackslash}p{0.04\linewidth}>{\centering\arraybackslash}p{0.05\linewidth}>{\centering\arraybackslash}p{0.035\linewidth}>{\centering\arraybackslash}p{0.05\linewidth}>{\centering\arraybackslash}p{0.035\linewidth}>{\centering\arraybackslash}p{0.035\linewidth}c>{\centering\arraybackslash}p{0.045\linewidth}>{\centering\arraybackslash}p{0.05\linewidth}>{\centering\arraybackslash}p{0.04\linewidth}>{\centering\arraybackslash}p{0.035\linewidth}}
        \toprule
         \textbf{Method} & \textbf{ARC\_c} & \textbf{ARC\_e} & \textbf{BoolQ}& \textbf{HellaS.}& \textbf{Lam.}& \textbf{OBQA}& \textbf{PIQA}& \textbf{SIQA}& \textbf{WinoG.}& \textbf{IFEval}& \textbf{MGSM} & \textbf{IncL.}&\textbf{Avg}\\
        \midrule
          BF16& 52.05& 80.3& 86.39& 58.41& 36.89& 31& 76.06& 43.09& 62.12&73.01& 58.1&70.28&60.64
\\
\midrule 
        \multicolumn{14}{c}{\textbf{INT8}}\\
         \midrule
          GPTQ+ds & \textbf{52.31}& 79.97& \textbf{86.38}& 58.32& 36.72& \textbf{31.1}& 76.25& \textbf{43.09}& \textbf{62.35}&74.45& \textbf{58.25}&69.73& 60.74
\\ \rowcolor{gray!15}
         \cellcolor{gray!20}GPTQ+qw & \cellcolor{gray!15}52.14 & \cellcolor{gray!15} \textbf{80.16} & \cellcolor{gray!15} 86.33 & \cellcolor{gray!15} \textbf{58.38} & \cellcolor{gray!15} \textbf{36.74} & \cellcolor{gray!15} 30.9 & \cellcolor{gray!15} \textbf{76.47} & \cellcolor{gray!15} 42.84 &  \cellcolor{gray!15}62 & \cellcolor{gray!15}\textbf{75.48} &  \cellcolor{gray!15}58.05 & \cellcolor{gray!15}\textbf{69.82} & \cellcolor{gray!15}\textbf{60.78}
\\
        GPTAQ+ds& 52.3& 80.09& \textbf{86.35}& \textbf{58.34}& 36.67& \textbf{31.2}& 76.2& \textbf{42.79}&61.64& 75.33& 53.93&\textbf{70.19}& 60.42
\\ 
\rowcolor{gray!15}
        \cellcolor{gray!20}GPTAQ+qw & \textbf{52.48}& \textbf{80.12}& 86.33& 58.24& \textbf{36.75}& 30.6& \textbf{76.34}& 42.48& \textbf{62.28}&\textbf{75.79}&\textbf{54.57}&70.18&\textbf{60.51}
\\
\midrule 
        \multicolumn{14}{c}{\textbf{INT4}}\\
        \midrule
         GPTQ+ds& 50.64& \textbf{78.18}& \textbf{84.46}& \textbf{56.68}& 35.36& \textbf{30}& 74.65& 41.05&\textbf{61.17}& 70.15&\textbf{51.72}&\textbf{65.05}&58.26
\\
\rowcolor{gray!15}
        \cellcolor{gray!20}GPTQ+qw & \textbf{51.07}& 77.76& 84.09& 56.24& \textbf{36.7}& 29.8& \textbf{74.95}& \textbf{41.74}& 59.87&\textbf{70.89}&51.37&64.77&\textbf{58.27}
\\
         AWQ+ds& \textbf{50.56}& 78.22&84.99& \textbf{57.05}& 35.45& \textbf{31.7}&\textbf{76.34}&\textbf{42.35}&60.74& 72.83& \textbf{51.92}&68.17& 59.19
\\
         \cellcolor{gray!20}AWQ+qw & \cellcolor{gray!15} 50.18 & \cellcolor{gray!15}\textbf{78.43} & \cellcolor{gray!15}\textbf{85.14} & \cellcolor{gray!15}56.94 & \cellcolor{gray!15}\textbf{36.68} & \cellcolor{gray!15}30.6 & \cellcolor{gray!15}75.85 & \cellcolor{gray!15}42.02 & \cellcolor{gray!15}\textbf{61.84} & \cellcolor{gray!15}\textbf{73.39} & \cellcolor{gray!15}51.25 & \cellcolor{gray!15}\textbf{68.9} & \cellcolor{gray!15}\textbf{59.27}
\\
         SPQR+ds& \textbf{51.84}&\textbf{78.26}& \textbf{84.65}&56.28& 34.36& \textbf{30.2}& 76.06&40.2&60.66&69.87&\textbf{49.78}&63.85&58.00
\\ 
\rowcolor{gray!15}
        \cellcolor{gray!20}SPQR+qw & 50.68 & 76.96 & 83.41 & \textbf{56.93} & \textbf{36.69} & 29.4 & \textbf{76.12} & \textbf{41.33} & \textbf{60.69} & \textbf{73.57} & 48.95 & \textbf{64.59} & \textbf{58.28}
\\
        GPTAQ+ds& 51.45 & 79.13 & \textbf{85.38} & \textbf{56.28} &35.42 & \textbf{30.7} & 75.22 & 41.3 & 61.41 & 71.86 & \textbf{53.52} & 66.06 & 58.98
\\ 
\rowcolor{gray!15}
        \cellcolor{gray!20}GPTAQ+qw & \textbf{52.09} & \textbf{79.15} & 85.23 & 56.12 & \textbf{37.35} & 30.6 & \textbf{76.55} & \textbf{42.15} & \textbf{62.47} & \textbf{73.78} & 53.1 & \textbf{67.25} & \textbf{59.65}
\\
\midrule 
        \multicolumn{14}{c}{\textbf{AVERAGE}}\\
         \midrule
          Quant+ds& \textbf{51.52}&\textbf{78.98}& \textbf{85.37}& \textbf{57.16}& 35.66& \textbf{30.82}& 75.79& 41.80& 61.33& 72.42& \textbf{53.19}&67.18& 59.27
\\ 
         \cellcolor{blue!10}Quant+qw & \cellcolor{blue!5}51.44 & \cellcolor{blue!5}78.76 & \cellcolor{blue!5}85.09 & \cellcolor{blue!5}57.14 & \cellcolor{blue!5}\textbf{36.82} & \cellcolor{blue!5}30.32 & \cellcolor{blue!5}\textbf{76.05} & \cellcolor{blue!5}\textbf{42.09} & \cellcolor{blue!5}\textbf{61.53} & \cellcolor{blue!5}\textbf{73.82} & \cellcolor{blue!5}52.88 & \cellcolor{blue!5}\textbf{67.59} & \cellcolor{blue!5}\textbf{59.46}\\
        \bottomrule
    \end{tabular}
    }
    \caption{Per-benchmark results on General tasks for DeepSeek-R1-0528-Qwen3-8B. Higher is better. }
    \label{tab:dpsk_r1_0528_qwen3_8b_gen}
\end{table*}

\begin{table*}[h]
    \centering
        {
    \small          

    \renewcommand{\arraystretch}{1.3} 
    \begin{tabular}{@{}cccccccc}
        \toprule
         \textbf{Bits} & \textbf{Method} & \textbf{AIME'24} & \textbf{AIME'25} & \textbf{Math-500}& \textbf{LiveCodeBench\_v5} & \textbf{LiveCodeBench\_v6} & \textbf{Avg}(\textuparrow)\\
          
        \midrule 
         BF16 & - & 86.67& 65.63& 93.6& 61.08& 50.86& 71.57
\\
        \midrule 
         \multirow{4}{*}{INT8} 
         & GPTQ 
         & 82.92 & 66.15 & 93.2 & \textbf{60.78} & 49.14 & 70.44
\\
         & \cellcolor{gray!20}+FAQ 
         & \cellcolor{gray!10}\textbf{84.17} & \cellcolor{gray!10}\textbf{68.65} & \cellcolor{gray!10}\textbf{93.5} & \cellcolor{gray!10}58.68 & \cellcolor{gray!10}\textbf{49.43} & \cellcolor{gray!10}\textbf{70.89}
\\
           & GPTAQ 
           & 82.78& 65.94& \textbf{92.93}& 58.08& \textbf{49.29}& 69.80
\\
\rowcolor{gray!20}
          \cellcolor{white!20}& \cellcolor{gray!20}+FAQ 
          & \textbf{83.06}& \textbf{67.09}& 92.67& \textbf{58.28}& 48.57& \textbf{69.93}
\\
        \midrule 
         \multirow{8}{*}{INT4} 
         & GPTQ 
         & 81.67& 62.61& \textbf{93.27}& 53.59& 44.57& 67.14
\\
\rowcolor{gray!15}
         \cellcolor{white!10}& \cellcolor{gray!20}+FAQ 
         & \textbf{83.06}& \textbf{63.20}& 93.20& \textbf{56.59}& \textbf{45.29}& \textbf{68.27}
\\
           & AWQ & 79.58& 63.12& 92.85& 53.44& 45.15& 66.83
\\
          & \cellcolor{gray!20}+FAQ & \cellcolor{gray!10}\textbf{81.46}& \cellcolor{gray!10}\textbf{64.9}& \cellcolor{gray!10}\textbf{93.7}& \cellcolor{gray!10}\textbf{55.54}& \cellcolor{gray!10}\textbf{45.43}& \cellcolor{gray!10}\textbf{68.21}
\\
           & SPQR 
           & 79.79& \textbf{61.15}& 91.9& 52.69& 40.29& 65.16
\\
\rowcolor{gray!15}
         \cellcolor{white!10}& \cellcolor{gray!20}+FAQ 
           & \textbf{81.25}& 60.22& \textbf{93.5}& \textbf{56.89}& \textbf{43.72}& \textbf{67.12}
\\
           & GPTAQ & \textbf{80.42}& 64.48& \textbf{93.2}& 54.94& 45.43& 67.69
\\
\rowcolor{gray!15}
         \cellcolor{white!10}& \cellcolor{gray!20}+FAQ & 80.83& \textbf{64.79}& 93.1& \textbf{55.84}& \textbf{45.57}& \textbf{68.03}
\\
         \midrule 
          \multirow{2}{*}{AVG} & Quant & 81.19& 63.91& 92.89& 55.59& 45.65& 67.85
\\
         & \cellcolor{blue!10}+FAQ & \cellcolor{blue!5}\textbf{82.31} & \cellcolor{blue!5}\textbf{64.81} & \cellcolor{blue!5}\textbf{93.28} & \cellcolor{blue!5}\textbf{56.97} & \cellcolor{blue!5}\textbf{46.34} & \cellcolor{blue!5}\textbf{68.74}\\
        \bottomrule
    \end{tabular}
    }
    \caption{Per-benchmark results on Specialized Domain tasks for DeepSeek-R1-0528-Qwen3-8B. Higher is better.}
    \label{tab:dpsk_r1_0528_qwen3_8b_special}
\end{table*}

\subsection{Detailed Experimental Setup}
\label{appendix:detailed_experimental_setup}
\subsubsection{Models and Quantization Methods.} Our investigation is conducted on a diverse set of recent open-source models with distinct architectures. Our model suite includes the standard dense Transformer Qwen3-8B; its reasoning-enhanced variant DeepSeek-R1-0528-Qwen3-8B, which is obtained by distilling chain-of-thought capabilities from DeepSeek-R1-0528 into the base model; and the larger-scale Qwen3-30B-A3B, a Mixture-of-Experts (MoE) model. This selection allows us to evaluate quantization performance across both dense and sparse architectures.
We benchmark four post-training quantization (PTQ) algorithms with our proposed FAQ: GPTQ, AWQ, SPQR, and GPTAQ. For GPTQ and GPTAQ, we explore both INT4 and INT8 precision levels, while AWQ and SPQR are evaluated in their standard INT4 configuration. Due to specific library or architectural constraints, particularly for the MoE model, Qwen3-30B-A3B is benchmarked using only the GPTQ and SPQR methods. For all quantization procedures, we consistently use a group\_size of 128 and a calibration set composed of 256 samples, as these are common and effective settings in the PTQ literature.

\subsubsection{Evaluation Benchmarks and Metrics.} We perform a comprehensive, multi-faceted evaluation to assess model performance post-quantization. All evaluations on downstream tasks are conducted under a strict zero-shot setting.
\begin{itemize}
    \item \textbf{Language Modeling}: We measure Perplexity (PPL) on Wikitext2, C4, and LAMBADA benchmarks.
    \item \textbf{General Reasoning and Multilingual Capabilities}: We evaluate performance on a broad suite of 12 downstream tasks: ARC-c and ARC-e, BoolQ, Hellaswag, LAMBADA, OpenBookQA, PIQA, SIQA, Winogrande, IFEval, MGSM and INCLUDE. For the benchmark IFEval, we report strict accuracy at the prompt level. For the multilingual math task MGSM, we use flexible-extract exact match. The INCLUDE benchmark, designed to evaluate regional knowledge, is tested specifically in its Chinese subset in our experiments.
    \item \textbf{Specialized Domain Capabilities}: To rigorously probe the limits of quantized models, we further tested them on highly challenging domain-specific benchmarks. For evaluating mathematical and logical reasoning skills,we employ high-level math benchmarks including professional exams AIME'24 and AIME'25, advanced mathematics MATH-500, and complex real-world coding challenges LiveCodeBench\_v5, LiveCodeBench\_v6.
\end{itemize}
For all accuracy-based evaluations on these tasks, we report the pass@1 metric, which means a single-attempt success rate.

\begin{table*}[h]
    \centering
        {
    \small          
    \setlength{\tabcolsep}{5 pt}
        
    \renewcommand{\arraystretch}{1.2}
    \begin{tabular}{@{}l>{\centering\arraybackslash}p{0.04\linewidth}c>{\centering\arraybackslash}p{0.04\linewidth}>{\centering\arraybackslash}p{0.05\linewidth}>{\centering\arraybackslash}p{0.035\linewidth}>{\centering\arraybackslash}p{0.05\linewidth}>{\centering\arraybackslash}p{0.035\linewidth}>{\centering\arraybackslash}p{0.035\linewidth}c>{\centering\arraybackslash}p{0.045\linewidth}>{\centering\arraybackslash}p{0.05\linewidth}>{\centering\arraybackslash}p{0.04\linewidth}>{\centering\arraybackslash}p{0.035\linewidth}}
        \toprule
         \textbf{Method} & \textbf{ARC\_c} & \textbf{ARC\_e} & \textbf{BoolQ}& \textbf{HellaS.}& \textbf{Lam.}& \textbf{OBQA}& \textbf{PIQA}& \textbf{SIQA}& \textbf{WinoG.}& \textbf{IFEval}& \textbf{MGSM} & \textbf{IncL.}&\textbf{Avg}\\
        \midrule
          BF16& 52.39& 79.50& 88.62& 59.61& 63.89& 34.40& 79.38& 43.14& 70.09&85.58& 61.57&84.04&66.85\\
\midrule 
        \multicolumn{14}{c}{\textbf{INT8}}\\
         \midrule
          GPTQ & 52.65 & 79.59 & 87.92 & 58.76 & 63.65 & 33.80 & 79.05 & 42.99 & 70.40 & \textbf{84.66} & 61.40 & 83.67 & 66.54
\\ \rowcolor{gray!15}
         \cellcolor{gray!20}
         +FAQ & \textbf{52.74} & \textbf{79.65} & \textbf{88.15} & \textbf{58.78} & \textbf{63.65} & \textbf{34.90} & \textbf{79.55} & \textbf{43.76} & \textbf{70.80} & 84.20 & \textbf{61.40} & \textbf{84.13} & \textbf{66.81}\\
\midrule 
        \multicolumn{14}{c}{\textbf{INT4}}\\
        \midrule
         GPTQ& \textbf{51.96} & 77.61 & 87.77 & 58.16 & \textbf{62.00} & \textbf{33.80} & 78.84 & \textbf{43.35} & 68.82 & 82.26 & \textbf{59.53} & 80.28 & 65.36
\\
\rowcolor{gray!15}
         \cellcolor{gray!20}
         +FAQ& 50.51 & \textbf{77.95} & \textbf{88.69} & \textbf{58.65} & 61.81 & 33.40 & \textbf{78.94} & 42.89 & \textbf{70.17} & \textbf{82.81} & 58.17 & \textbf{82.39} & \textbf{65.53}
\\
         SPQR& 50.77 & 78.03 & 87.83 & \textbf{58.02} & \textbf{61.42} & \textbf{33.60} & 78.40 & 42.48 & 68.51 & 82.44 & 60.03 & 79.63 & 65.10
\\ 
\rowcolor{gray!15}
         \cellcolor{gray!20}
         +FAQ& \textbf{51.54} & \textbf{78.70} & \textbf{87.83} & 57.88 & 60.26 & 32.40 & \textbf{78.67} & \textbf{42.84} & \textbf{68.90} & \textbf{83.55} & \textbf{60.43} & \textbf{81.28} & \textbf{65.36} \\
\midrule 
        \multicolumn{14}{c}{\textbf{AVERAGE}}\\
         \midrule
          Quant& \textbf{51.79}&78.41& 87.84& 58.31& \textbf{62.36}& \textbf{33.73}& 78.76& 42.94& 69.24& 83.12& \textbf{60.32}&81.19& 65.67
\\ \rowcolor{blue!5}
         \cellcolor{blue!10}
         +FAQ& 51.60& \textbf{78.77}& \textbf{88.22}& \textbf{58.44}& 61.91& 33.57& \textbf{79.05}& \textbf{43.16}& \textbf{69.96}& \textbf{83.52}& 60.00& \textbf{82.60}& \textbf{65.90} \\
        \bottomrule
    \end{tabular}
    }
    \caption{Per-benchmark results on General tasks for Qwen3-30B-A3B. Higher is better. }
    \label{tab:qwen3_30b_a3b_gen}
\end{table*}

\subsection{Detailed Experimental Results}

This section provides the detailed, per-benchmark results for the experiments discussed in Subsections 4.3, Ablation Studies and Generalization, of the main paper. Due to space constraints, only the averaged results were presented in the main body. The following tables offer a granular view of the performance across all evaluated tasks, providing the full empirical evidence for our conclusions.

\subsubsection{Detailed Results for the "Family-Aware" Hypothesis Validation}

The following Table~\ref{tab:dpsk_r1_0528_qwen3_8b_gen} and \ref{tab:dpsk_r1_0528_qwen3_8b_special} detail the per-benchmark performance comparison of Family-Sourced (Qwen3-235B, denoted as `+qw`) versus Knowledge-Sourced (DeepSeek-R1, denoted as `+ds`) calibration data on the distilled DeepSeek-R1-0528-Qwen3-8B model.

\subsubsection{Detailed Results for Generalization to MoE Architectures}
The following Table~\ref{tab:qwen3_30b_a3b_gen} and \ref{tab:qwen3_30b_a3b_special} present the granular, per-benchmark results for the Qwen3-30B-A3B MoE model, comparing baseline quantization methods against their FAQ-enhanced counterparts (`+FAQ`).

\begin{table*}[h]
    \centering
        {
    \small          

    \renewcommand{\arraystretch}{1.3} 
    \begin{tabular}{@{}cccccccc}
        \toprule
         \textbf{Bits} & \textbf{Method} & \textbf{AIME'24} & \textbf{AIME'25} & \textbf{Math-500}& \textbf{LiveCodeBench\_v5} & \textbf{LiveCodeBench\_v6} & \textbf{Avg}(\textuparrow)\\
          
        \midrule 
         BF16 & - & 81.67
& 71.88
& 95.40& 63.47
& 59.43
& 74.37
\\
        \midrule 
         \multirow{2}{*}{INT8} 
         & GPTQ 
         & 79.72 & 68.06 & 94.85& 59.28 & \textbf{55.58} & 71.50
\\
         & \cellcolor{gray!25}+FAQ 
         & \cellcolor{gray!10}\textbf{80.56} & \cellcolor{gray!10}\textbf{70.11} & \cellcolor{gray!10}\textbf{94.85}& \cellcolor{gray!10}\textbf{60.48} & \cellcolor{gray!10}55.43 & \cellcolor{gray!10}\textbf{72.28}
\\
        \midrule 
         \multirow{4}{*}{INT4} 
         & GPTQ 
         & 79.38 & 69.17 & \textbf{94.75} & \textbf{60.63} & 53.72 & 71.53
\\
        & \cellcolor{gray!25}+FAQ 
         & \cellcolor{gray!10}\textbf{79.79} & \cellcolor{gray!10}\textbf{70.21} & \cellcolor{gray!10}94.68 & \cellcolor{gray!10}59.88 & \cellcolor{gray!10}\textbf{54.36} & \cellcolor{gray!10}\textbf{71.78}
\\
           & SPQR 
           & \textbf{80.32} & 68.23 & 94.70 & 57.78 & 52.43 & 70.69
\\
\rowcolor{gray!15}
         \cellcolor{white!10}& \cellcolor{gray!25}+FAQ 
           & 80.00 & \textbf{68.44} & \textbf{95.50} & \textbf{61.08} & \textbf{53.29} & \textbf{71.66}\\
         \midrule 
          \multirow{2}{*}{AVG} & Quant & 79.81& 68.49& 94.77& 59.23& 53.91& 71.24
\\
         & \cellcolor{blue!10}+FAQ & \cellcolor{blue!5}\textbf{80.12}& \cellcolor{blue!5}\textbf{69.59}& \cellcolor{blue!5}\textbf{95.01}& \cellcolor{blue!5}\textbf{60.48}& \cellcolor{blue!5}\textbf{54.36}& \cellcolor{blue!5}\textbf{71.91}\\
        \bottomrule
    \end{tabular}
    }
    \caption{Per-benchmark results on Specialized Domain tasks for Qwen3-30B-A3B. Higher is better.}
    \label{tab:qwen3_30b_a3b_special}
\end{table*}

\subsection{Detailed Implementation} 

Our evaluation pipeline is built on the lm-evaluation-harness and EvalScope toolkits to ensure standardized and reproducible assessment, as illustrated in code Listing~\ref{lst:loglikelihood_rolling},\ref{lst:generate_until} and \ref{lst:evalscope_sh} . High-performance inference is enabled by the SGLang and vLLM serving engines in code Listing~\ref{lst:sglang} and . To guarantee the statistical robustness and reliability of our findings, we adopt a stringent evaluation protocol. Each model-method configuration is evaluated 4 to 8 times on perplexity and general accuracy benchmarks, with the average score being reported. Recognizing the inherent high variance in complex reasoning tasks, we significantly increase the evaluation runs for the AIME datasets to a range of 32 to 64, reporting the averaged accuracy to provide a highly stable and credible performance measure. All experiments were performed on NVIDIA H20 GPUs.

\subsubsection{Testing and Model Deployment Commands.}

In the experiments of this paper, we used lm\_eval\footnote{https://github.com/EleutherAI/lm-evaluation-harness} for evaluating the perplexity and general datasets, and evalscope\footnote{https://github.com/modelscope/evalscope} for evaluating the special dataset. The script commands used during the evaluations are shown in code listing~\ref{lst:loglikelihood_rolling}, \ref{lst:generate_until}, and \ref{lst:evalscope_sh}. When using lm\_eval for evaluation, it is necessary to distinguish between the loglikelihood\_rolling mode and the generate\_until mode. Specifically, the generate\_until mode is used for the IFEval and MGSM datasets, while the loglikelihood\_rolling mode is used for the others.

To complete the dataset evaluation experiments faster, better, and more efficiently, we chose the mainstream inference service engines SGLang\footnote{https://github.com/sgl-project/sglang} and vLLM\footnote{https://github.com/vllm-project/vllm} for model deployment and inference. The corresponding model deployment command codes are shown in listing~\ref{lst:sglang} and \ref{lst:vllm}. It is important to note that due to the characteristics of the IFEval dataset evaluation, only vLLM can be used for deployment and API calls during testing.

Throughout the entire experiment, the versions of the relevant Python libraries used are as follows: torch 2.7.0, transformers 4.53.0, gptqmodel 4.0.0, sglang 0.4.9, vllm 0.9.1, lm\_eval 0.4.9, and evalscope 2.0.0 .

\subsection{Additional Experiments on Model Generalization and Data Generation Strategies} 

In this section, we present supplementary experiments designed to further explore the generalization capabilities and underlying mechanisms of our proposed method, FAQ.

\subsubsection{Generalization Across Different Model Families}

To validate that the effectiveness of FAQ extends beyond the Qwen model family, we conducted evaluations on two additional, widely-used open-source model series: Llama3 and DeepSeek.

\begin{table*}[htbp!]
    \centering
        {
    \small          

            \setlength{\tabcolsep}{5 pt}
    \renewcommand{\arraystretch}{1.2}
    \begin{tabular}{@{}l>{\centering\arraybackslash}p{0.04\linewidth}c>{\centering\arraybackslash}p{0.04\linewidth}>{\centering\arraybackslash}p{0.05\linewidth}>{\centering\arraybackslash}p{0.035\linewidth}>{\centering\arraybackslash}p{0.05\linewidth}>{\centering\arraybackslash}p{0.035\linewidth}>{\centering\arraybackslash}p{0.035\linewidth}c>{\centering\arraybackslash}p{0.045\linewidth}>{\centering\arraybackslash}p{0.05\linewidth}>{\centering\arraybackslash}p{0.04\linewidth}>{\centering\arraybackslash}p{0.035\linewidth}}
        \toprule
         \textbf{Method} & \textbf{ARC\_c} & \textbf{ARC\_e} & \textbf{BoolQ}& \textbf{HellaS.}& \textbf{Lam.}& \textbf{OBQA}& \textbf{PIQA}& \textbf{SIQA}& \textbf{WinoG.}& \textbf{IFEval}& \textbf{MGSM} & \textbf{IncL.}&\textbf{Avg}\\
        \midrule
          BF16& 53.67& 82.24& 85.38& 59.78& 78.38& 36.20& 80.30& 42.32& 70.40&73.75& 59.30&54.68&64.70\\
\midrule 
        \multicolumn{14}{c}{\textbf{INT8}}\\
         \midrule
          GPTQ & 53.67& 82.49& 85.32& 59.76& 78.34& 35.80& 80.20& 42.37& 70.48& 73.38& 58.50& 54.50& 64.57\\ \rowcolor{gray!15}
         \cellcolor{gray!20}
         +FAQ & 53.88& 82.45& 85.23& 59.81& 78.38& 35.50& 80.15& 42.20& 70.32& 74.40& 58.82& 55.05& 64.68\\
\midrule 
        \multicolumn{14}{c}{\textbf{INT4}}\\
        \midrule
         GPTQ& 50.17& 81.14& 83.76& 58.61& 77.78& 33.80& 79.49& 41.66& 69.61& 70.06& 55.77& 52.66& 62.88\\
\rowcolor{gray!15}
         \cellcolor{gray!20}
         +FAQ& 49.49& 80.72& 84.68& 58.97& 77.06& 33.20& 79.16& 41.97& 70.40& 71.35& 57.70& 51.56& 63.02\\
         GPTAQ& 50.60& 80.56& 84.34& 58.83& 77.97& 34.00& 79.43& 40.38& 70.01& 71.90& 56.60& 50.28& 62.91\\ 
\rowcolor{gray!15}
         \cellcolor{gray!20}
         +FAQ& 49.74& 80.62& 83.54& 58.85& 76.91& 35.40& 79.03& 41.94& 69.42& 73.11& 56.22& 53.67& 63.20\\
\midrule 
        \multicolumn{14}{c}{\textbf{AVERAGE}}\\
         \midrule
          Quant& 51.48&81.40& 84.47& 59.07& 78.03& 34.53& 79.71& 41.47& 70.03& 71.78& 56.96&52.48& 63.45
\\ \rowcolor{blue!5}
         \cellcolor{blue!10}
         +FAQ& 51.04& 81.26& 84.48& 59.21& 77.45& 34.70& 79.45& 42.04& 70.05& 72.95& 57.58& 53.43& 63.64\\
        \bottomrule
    \end{tabular}
    }
    \caption{Per-benchmark results on General tasks for Llama3.1-8B-Instruct. Higher is better. }
    \label{tab:llama3.1_8b_gen}
\end{table*}

\begin{table*}[h]
    \centering
        {
    \small          

    \renewcommand{\arraystretch}{1.3} 
    \begin{tabular}{@{}cccccc}
        \toprule
         \textbf{Model}& \textbf{Method} & \textbf{AIME'24} & \textbf{Math-500}& \textbf{LiveCodeBench\_v5} & \textbf{Avg}(\textuparrow)\\
        \midrule 
         \multirow{3}{*}{DeepSeek-R1}& BF16& 86.25& 95.73& 72.75& 84.91\\
         & w8a8-int8& 81.67& 95.38& 71.26& 82.77\\
 & \cellcolor{gray!25}+FAQ& \cellcolor{gray!10}83.47& \cellcolor{gray!10}95.69& \cellcolor{gray!10}72.16&\cellcolor{gray!10}\textbf{83.77}\\
        \midrule 
         \multirow{4}{*}{Qwen3-8B}& BF16& 83.33& 95.20& 58.08& 78.87\\
        & w8a8-int8& 74.33& 94.33& 54.67& 74.44\\

         & \cellcolor{gray!25}+FAQ-8B& 75.0& 94.60& 55.09& 74.90\\
 \rowcolor{gray!15}\cellcolor{white!10}& \cellcolor{gray!25}+FAQ-235B& 76.36& 94.53& 55.61&\cellcolor{gray!10}\textbf{75.5}\\
    \bottomrule
    \end{tabular}
    }
    \caption{Evaluation of INT8 quantization on challenging domain-specific benchmarks: AIME'24, Math-500, and LiveCodeBench (coding). This table serves two purposes: (1) It demonstrates the effectiveness of our FAQ method on the DeepSeek-R1 model. (2) It presents an ablation study on Qwen3-8B, comparing the impact of using data generated by the model itself (+FAQ-8B) versus a larger "Elder Sibling," Qwen3-235B (+FAQ-235B). The superior performance of +FAQ-235B highlights the benefit of using a more capable data generator. All values are pass@1 accuracy (\%). Higher is better.}
    \label{tab:self_special}
\end{table*}

\begin{itemize}
    \item \textbf{Models and Data Generation}: We evaluated Llama3.1-8B-Instruct and the DeepSeek-R1 model. Consistent with our main methodology, we employed a more capable "teacher" model for synthetic data generation where available. For Llama3.1-8B-Instruct, data was generated by the Llama3-405B model. For DeepSeek-R1, where a significantly larger public model is unavailable, we adopted a self-generation strategy.
    \item \textbf{Results and Analysis}: The results for Llama3.1-8B-Instruct are presented in Table \ref{tab:llama3.1_8b_gen}. Our FAQ method consistently enhances performance, particularly in the challenging INT4 setting. For example, applying FAQ to GPTAQ (GPTAQ+FAQ) improves the average accuracy to 63.20\%, surpassing both the standard GPTQ baseline (62.88\%) and the standalone GPTAQ (62.91\%). This demonstrates that FAQ's performance benefits are robust and generalizable across different foundational model architectures.
\end{itemize}

\begin{figure*}[t]
\centering
\includegraphics[width=2\columnwidth]{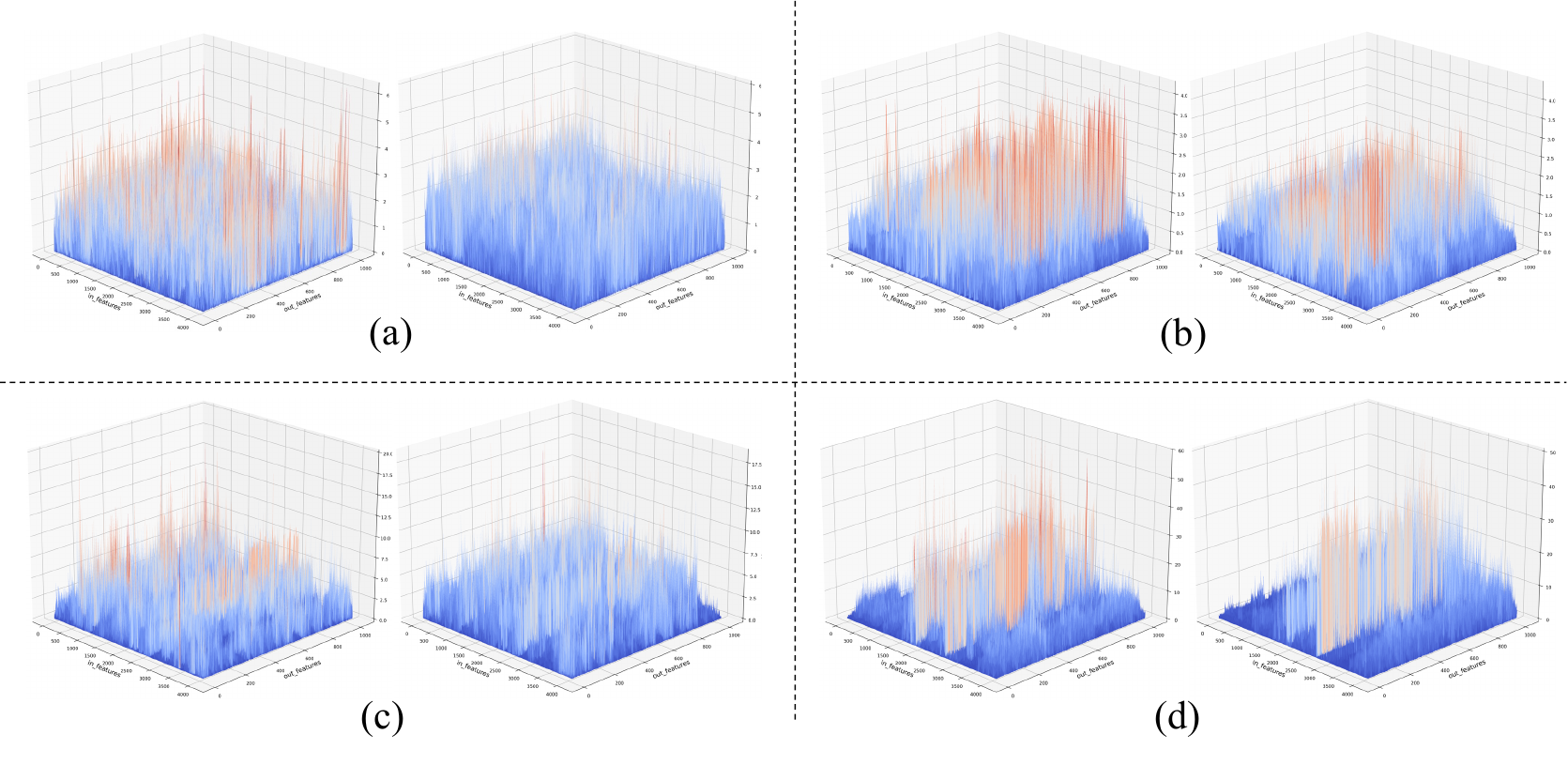}
\caption{ Visualization of activation distributions, demonstrating the outlier suppression effect of FAQ. The subplots (a)-(d)  displays the activation value distributions at the input of the self-attention output projection ($O_{proj}$) within the Qwen3-8B model.  Each pair compares the activations induced by the baseline calibration data (\textbf{left}) versus our FAQ-generated data (\textbf{right}).  The FAQ-generated data consistently produces a smoother activation landscape with substantially suppressed outliers (fewer and shorter red peaks), directly illustrating its effectiveness in creating a more quantization-amenable distribution.}
\label{3d}
\end{figure*}

\begin{figure*}[t]
  \centering
  \begin{subfigure}[t]{0.32\textwidth}
    \centering
    \includegraphics[width=\linewidth]{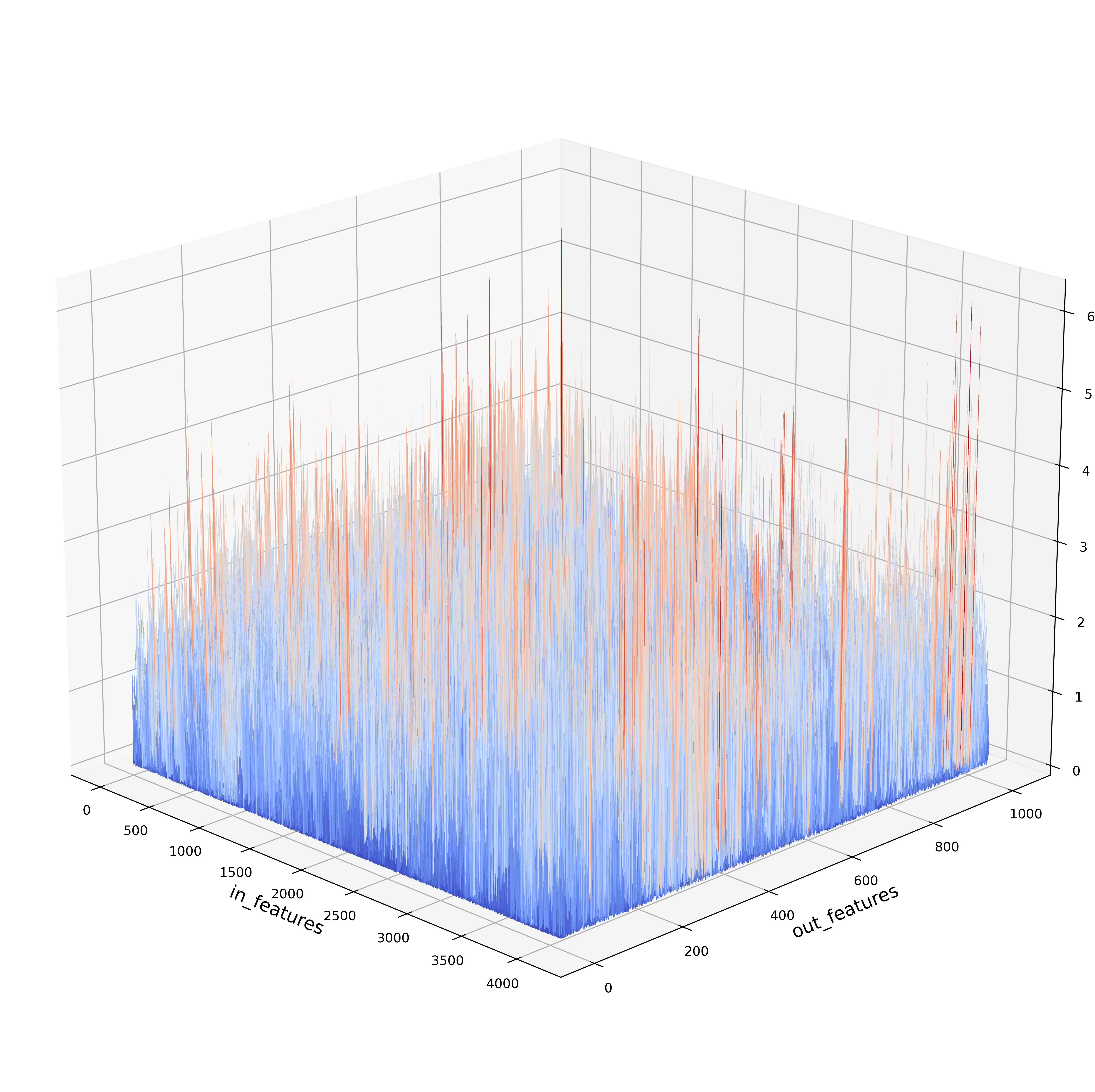}
    \caption{Seed calibration data}
    \label{fig:act_seed}
  \end{subfigure}\hfill
  \begin{subfigure}[t]{0.32\textwidth}
    \centering
    \includegraphics[width=\linewidth]{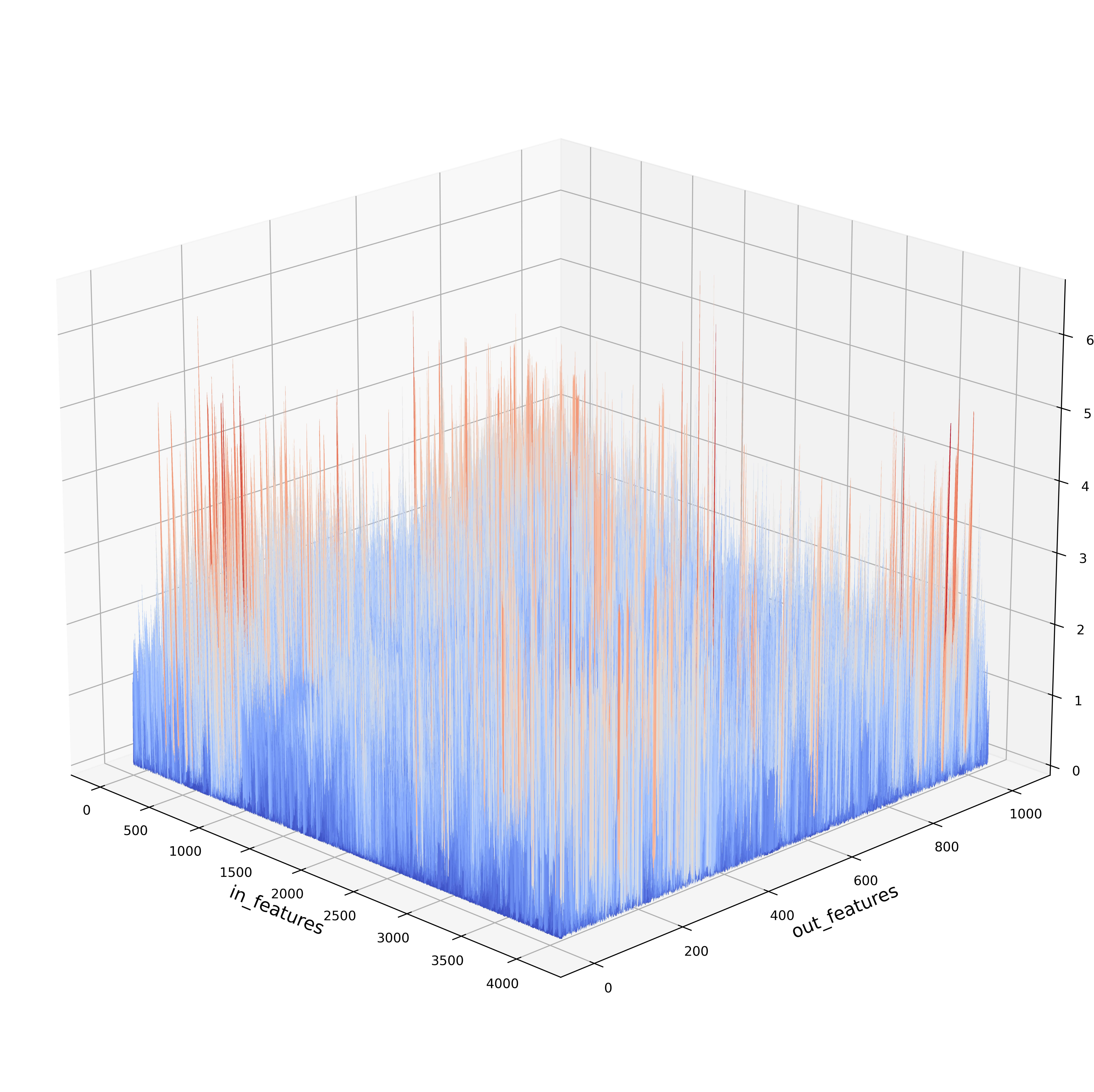}
    \caption{Regenerated by Qwen3-8B}
    \label{fig:act_self_8b}
  \end{subfigure}\hfill
  \begin{subfigure}[t]{0.32\textwidth}
    \centering
    \includegraphics[width=\linewidth]{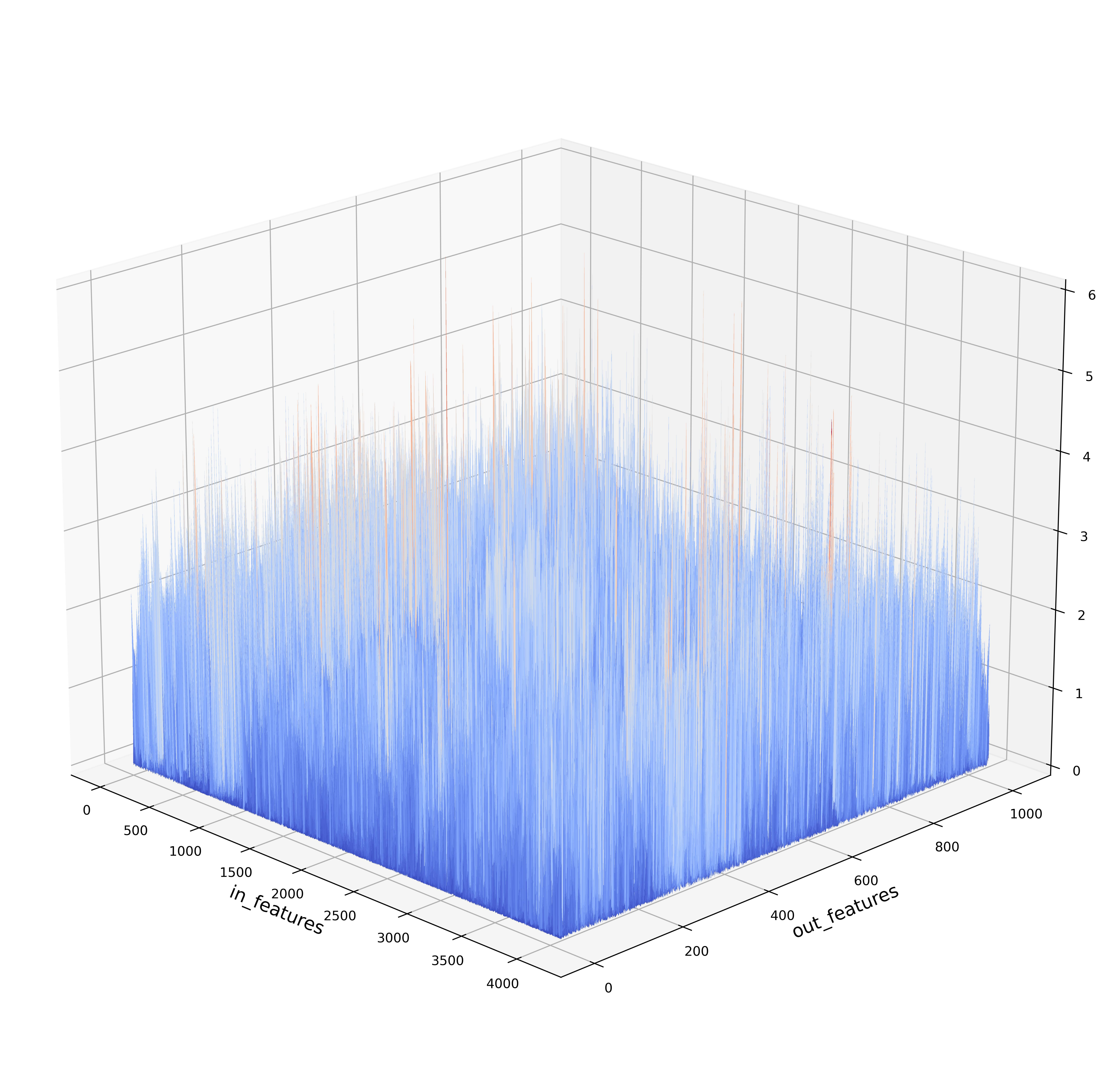}
    \caption{Regenerated by Qwen3-235B}
    \label{fig:act_faq_235b}
  \end{subfigure}
  \caption{Why a larger in-family generator yields better calibration for PTQ.
We plot the activation landscape of the \emph{same} quantized target model (Qwen3-8B) under an identical PTQ configuration, measured at the Layer-23 self-attention output projection (O-proj). We vary only the calibration data used for calibration: (a) original seed data, (b) data regenerated by the target model itself (Qwen3-8B; self), and (c) data regenerated by a larger in-family model (Qwen3-235B; FAQ). Compared to (a) and (b), the elder-sibling generator in (c) induces a noticeably smoother landscape with substantially fewer and lower outlier spikes, supporting the choice of a more capable in-family model for calibration regeneration.}

  \label{fig:three_acts}
\end{figure*}

\subsubsection{Ablation Study on the Impact of Data Generator's Model Size}

The central premise of our method, FAQ, is that leveraging high-quality synthetic data is crucial for achieving robust post-training quantization. To provide strong empirical support for this premise and to specifically investigate the role of the data generator's capability, we conducted a critical ablation study on the Qwen3-8B model. This study directly compares the quantization outcomes when using data from two distinct sources: a significantly larger "Elder Sibling" model (the Qwen3-235B), versus the model's own self-generated data. The goal is to demonstrate that the quality of the data source directly correlates with the final performance of the quantized model.

The results of this ablation, presented in Table \ref{tab:self_special}, unequivocally support the core principle of FAQ. While the self-generation strategy already offers a substantial improvement over baseline quantization methods, leveraging data from the more capable "Elder Sibling" model consistently yields superior results, more effectively closing the performance gap to the unquantized BF16 baseline. This finding provides direct validation for our method's core idea: the quality and capability of the data generator are key determinants of post-quantization robustness. It also highlights the practical versatility of FAQ, as it remains a highly effective framework even when an "Elder Sibling" model is unavailable and a self-generation strategy must be employed.

\begin{figure*}[htbp]
\begin{lstlisting}[
    caption={lm\_eval:loglikelihood\_rolling}, 
    label=lst:loglikelihood_rolling, 
    % language=bash 
]
# wikitext,c4
lm_eval --tasks wikitext,c4 \
  --model local-completions \
  --model_args max_length=16384,model=${model_name},base_url=http://${model_ip}:${model_port}/v1/completions,num_concurrent=32,max_retries=3,tokenized_requests=False,timeout=72000 \
  --num_fewshot 0 \
  --gen_kwargs temperature=0.6,max_gen_toks=32768,top_k=20,top_p=0.95 

\end{lstlisting}
\end{figure*}

\begin{figure*}[htbp]
\begin{lstlisting}[
    caption={lm\_eval:generate\_until}, 
    label=lst:generate_until, 
    % language=bash 
]
# ifeval
lm_eval --tasks ifeval \
  --model local-chat-completions \
  --model_args max_length=32768,model=${model_name},base_url=http://${model_ip}:${model_port}/v1/chat/completions,num_concurrent=32,max_retries=3,tokenized_requests=False,timeout=72000 \
  --num_fewshot 0 \
  --apply_chat_template qwen3 \
  --gen_kwargs temperature=0.6,max_gen_toks=32768,top_k=20,top_p=0.95 
\end{lstlisting}
\end{figure*}

\begin{figure*}[htbp!]
\begin{lstlisting}[
    caption={evalscope}, 
    label=lst:evalscope_sh, 
]

# AIME2024,Aime2025,MATH500,LiveCodeBench
evalscope eval --datasets aime24 aime25 math_500 live_code_bench \
  --model ${model_name} \
  --api-url http://${host}:${port}/v1/chat/completions \
  --api-key EMPTY \
  --eval-type service \
  --eval-batch-size 4 \
  --generation-config '{"do_sample":true,"temperature":0.6,"top_p":0.95,"top_k":20,"max_tokens":32768,"n":8,"chat_template_kwargs":{"enable_thinking": true}}' \
  --chat-template qwen3 \
  --stream \

\end{lstlisting}
\end{figure*}

\begin{figure*}[htbp!]
\begin{lstlisting}[
    caption={sglang}, 
    label=lst:sglang, 
]
python -m sglang.launch_server \
    --model-path models/Qwen3-30B-A3B --trust-remote-code \
    --served-model-name Qwen3-30B-A3B --port ${port} \
    --tensor-parallel-size 4 --mem-fraction-static 0.9 --attention-backend fa3 \
    --reasoning-parser qwen3 --enable-torch-compile \
    --torch-compile-max-bs 32 --cuda-graph-max-bs 32
    

\end{lstlisting}
\end{figure*}

\begin{figure*}[htbp!]
\begin{lstlisting}[
    caption={vllm}, 
    label=lst:vllm, 
]
python3 -m vllm.entrypoints.openai.api_server \
    --model /models/Qwen3-8B --trust-remote-code \
    --served-model-name Qwen3-8B --port ${port} \
    --tensor-parallel-size 4 --gpu-memory-utilization 0.85 \
    --max-model-len 34816 --reasoning-parser qwen3 \
    --max-num-seqs 32
    
    
\end{lstlisting}
\end{figure*}

\end{document}